\documentclass[12pt]{article}
\usepackage{amsmath,amsthm,amssymb,bm,mathrsfs}
\usepackage{graphicx}
\usepackage[dvipsnames]{xcolor}
\usepackage{bibunits}
\usepackage[ruled,linesnumbered]{algorithm2e}

\usepackage{caption,enumerate,listings}
\usepackage{setspace}
\usepackage[OT1]{fontenc}
\usepackage{natbib}

\usepackage[hidelinks]{hyperref}
\usepackage{pifont}
\usepackage[margin=1in]{geometry}
\usepackage[protrusion=false,expansion=true]{microtype}
\usepackage[title]{appendix}
\usepackage{bbm}
\usepackage{subcaption}
\usepackage{comment}

\usepackage{enumitem}
\usepackage{multirow}
\usepackage{titletoc}
\newcommand{\argmin}{\mathop{\mathrm{argmin}}}

\newtheorem{proposition}{{\bf Proposition}}
\newtheorem{corollary}{{\bf Corollary}}
\newtheorem{theorem}{{\bf Theorem}}
\newtheorem{assumption}{{\bf Assumption}}
\newtheorem{definition}{{\bf Definition}}
\newtheorem{remark}{{\bf Remark}}
\usepackage{setspace}
\usepackage{booktabs}
\theoremstyle{plain}
\usepackage{algorithm}
\usepackage{algorithmic}
\usepackage{titlesec}
\usepackage{tocloft}
\titleformat{\section}[block]{\normalfont\large\bfseries}{\thesection}{1em}{}
\makeatletter
\g@addto@macro\normalsize{%
  \setlength{\abovedisplayskip}{5pt}%
  \setlength{\belowdisplayskip}{5pt}%
  \setlength{\abovedisplayshortskip}{5pt}%
  \setlength{\belowdisplayshortskip}{5pt}%
}

\floatname{algorithm}{Fresh Data Augmentation Case}

\title{\Large \bf Recursive Learning Without Collapse: A Weighting-Based Stabilization Framework}
\author{ Hengzhi He$^{\S}$\footnote{The first two authors contribute equally to this work.} \quad Shirong Xu$^{\ddagger *}$\footnote{Address for correspondence: Shirong Xu, Wang Yanan Institute for Studies in Economics, Xiamen University, China. Email: shirongxu5566@xmu.edu.cn } \quad Guang Cheng$^\S$ \\ $^\S$Department of Statistics and Data Science, \\ University of California, Los Angeles \\ $\ddagger$ Wang Yanan Institute for Studies in Economics, \\ Xiamen University }

\date{To appear in Journal of the Royal Statistical Society: Series B.
}

\begin{document}

\maketitle

\begin{abstract}
Recent studies identified an intriguing phenomenon in recursive generative model training known as model collapse, where models trained on data generated by previous models exhibit severe performance degradation. Addressing this issue and developing more effective training strategies have become central challenges in generative model research. In this paper, we investigate this phenomenon within a novel framework, where generative models are iteratively trained on a combination of newly collected real data and synthetic data from the previous training step. To develop an optimal training strategy for integrating real and synthetic data, we evaluate the performance of a weighted training scheme in various scenarios, including Gaussian distribution estimation, generalized linear models, and nonparametric estimation. We theoretically characterize the impact of the mixing proportion and weighting scheme of synthetic data on the final model's performance. Our key finding is that, across different settings, the optimal weighting scheme under different proportions of synthetic data asymptotically follows a unified expression, revealing a fundamental trade-off between leveraging synthetic data and model performance. In some cases, the optimal weight assigned to real data corresponds to the reciprocal of the golden ratio. Finally, we validate our theoretical results on extensive simulated datasets and a real tabular dataset.
\end{abstract}

\begin{bibunit}[apalike]
%\setstretch{1}
 
\setstretch{1.7}
\section{Introduction}

In recent years, synthetic data have been widely used to train generative models, especially in the domain of large language models \citep{meng2022generating,shumailov2023curse,dohmatob2024tale} and computer vision \citep{mishra2022task2sim}. This trend is mainly motivated by the limited availability of data to train larger models due to neural scaling laws \citep{bansal2022systematic,villalobos2024will}. However, several critical issues arise regarding the utility of synthetic data for training purposes. For instance, because synthetic data do not perfectly align with the real data distribution, models trained on synthetic data are prone to have degraded performance. This phenomenon has been extensively validated in the literature \citep{xu2023utility,wong2016understanding,dohmatob2024strong}.

\begin{figure}[h]
    \centering
    \includegraphics[scale=0.26]{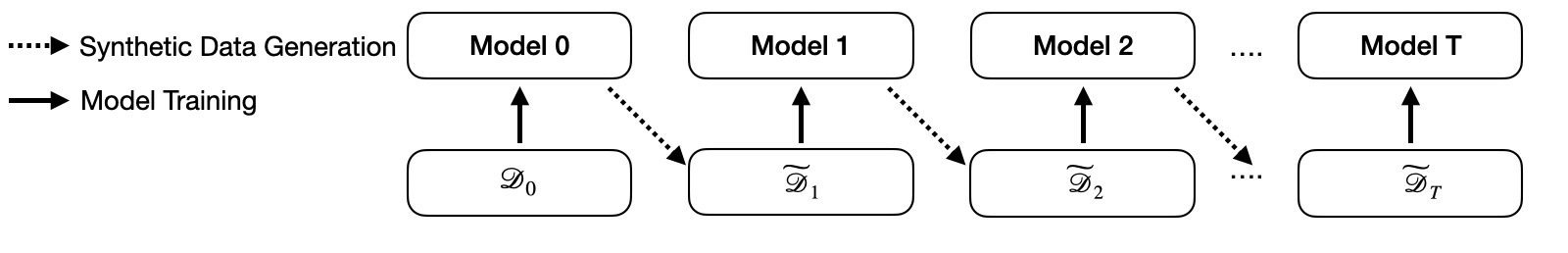}
    \caption{The general framework of model collapse phenomenon during recursive training \citep{shumailov2024ai}. In this framework, $\mathcal{D}_0$ denotes the initial real dataset and $\widetilde{\mathcal D}_{t}$ denotes the synthetic dataset generated by the $(t-1)$-th generative model, which is then used to train the $t$-th generative model.}
    \label{fig:ModelCollapse_1}
\end{figure}

A recent study by \citet{shumailov2024ai} reveals that AI models may collapse when trained on recursively generated data. The recursive training framework is illustrated in Figure \ref{fig:ModelCollapse_1}, where generative models are trained on synthetic data produced by earlier generative models. Over successive training iterations, these models gradually lose information about the real data distribution, a phenomenon known as \textit{model collapse}. For example, \citet{shumailov2024ai} use Gaussian distribution estimation to demonstrate how repeated estimation over iterations causes the estimated covariance matrix to almost surely collapse to zero, while the sample mean diverges. Similar results in a linear regression context are validated by \citet{dohmatob2024model}. In this paper, we refer to this framework (illustrated in Figure~\ref{fig:ModelCollapse_1}) as the \textit{fully synthetic} framework.

\begin{figure}[h]
    \centering
    \includegraphics[scale=0.26]{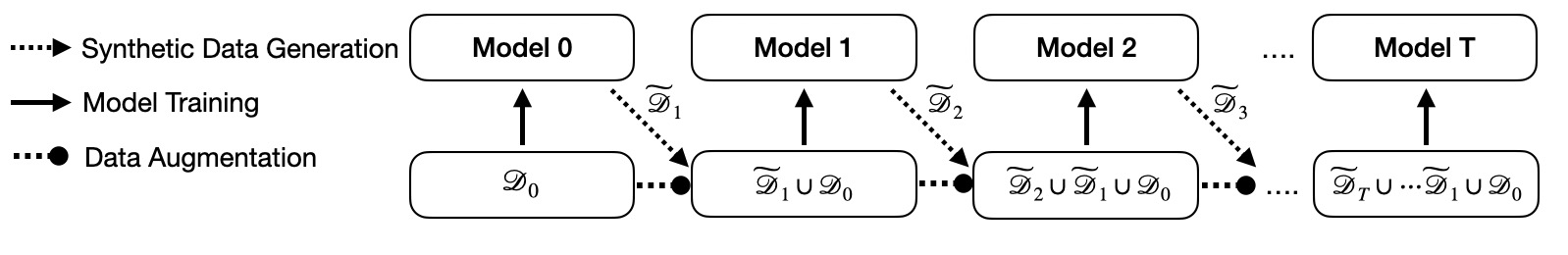}
    \caption{The general framework for mitigating the model collapse phenomenon during recursive training involves accumulating data \citep{gerstgrasser2024model, kazdan2024collapse}. In this framework, the training dataset is progressively expanded at each step $t$ by incorporating synthetic data from last generative model.}
    \label{fig:ModelCollapse_2}
\end{figure}

Subsequently, \cite{gerstgrasser2024model} and \cite{kazdan2024collapse} explore model collapse in a new scenario, illustrated in Figure~\ref{fig:ModelCollapse_2}, in which a model is trained iteratively, with each training step leveraging the original real training dataset $\mathcal{D}_0$ and all synthetic data in previous steps. Theoretically, they demonstrate that in both regression and Gaussian distribution estimation settings, the phenomenon of model collapse is avoided, showing that the expected test error remains bounded even as the number of iterations increases. In other words, model collapse is prevented by integrating all the data from previous steps into the training process. Furthermore, they demonstrate through experiments that data accumulation effectively prevents model collapse across various models, including LLMs and diffusion models. Later, such a data accumulation scheme is proved to be effective when the underlying generative models belong to exponential families \citep{dey2024universality}. In this paper, we refer to this framework (illustrated in Figure~\ref{fig:ModelCollapse_2}) as the \textit{synthetic accumulation} framework.

%\cite{dey2024universality} extends the theoretical results of \cite{gerstgrasser2024model} and \cite{kazdan2024collapse}, showing that model collapse can be avoided by data accumulation\ during training when the underlying generative models belong to exponential families. In this paper, we refer to this framework (illustrated in Figure~\ref{fig:ModelCollapse_2}) as the \textit{synthetic accumulation} framework.

%For instance, Model 2 is trained on a dataset that augments $\mathcal{D}_0$ with synthetic datasets $\widetilde{D}_{1}$ and $\widetilde{D}_{2}$ generated by Model 0 and Model 1, respectively. 

%argue that the fully synthetic setting depicted in Figure~\ref{fig:ModelCollapse_1}—where the training data for each iteration is entirely generated by the previous iteration—is unrealistic. Instead, they consider 

%Their theoretical findings are empirically validated on image datasets using different diffusion models, demonstrating that the final generative model tends to exhibit greater stability when trained with a smaller proportion of synthetic data. 

\begin{figure}[h!]
    \centering
    \includegraphics[scale=0.26]{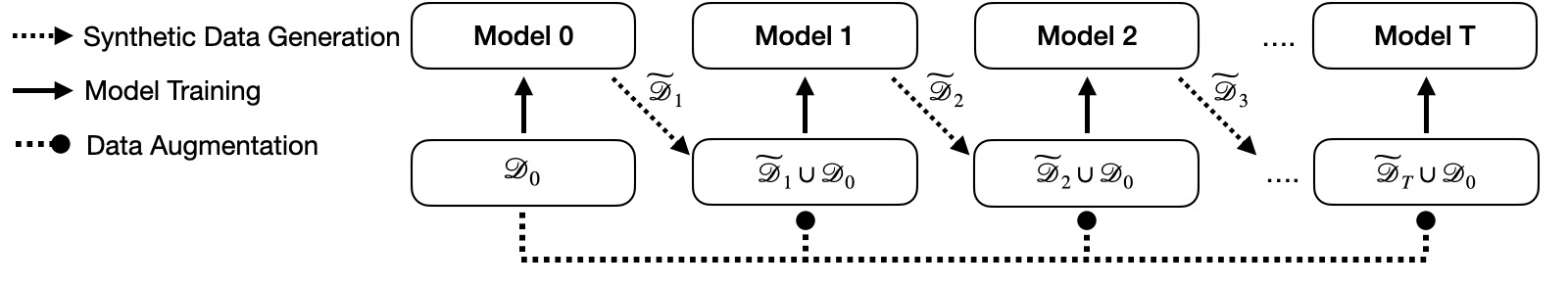}
    \caption{The general framework for avoiding the model collapse phenomenon during recursive training by augmenting only real data \citep{bertrand2024on}. In this framework, at the $t$-th training step, the generative model is trained based on $\widetilde{\mathcal D}_{t} \cup  \mathcal{D}_0$.
}
    \label{fig:ModelCollapse_3}
\end{figure}

A key result of the synthetic accumulation framework is that it incorporates {\em excessive} synthetic data during training, leading to poorer performance in the final model compared to one trained purely on real data \citep{kazdan2024collapse}. In contrast, \citet{bertrand2024on} consider a data augmentation approach that utilizes only synthetic data generated by the most recent generative model, as shown in Figure~\ref{fig:ModelCollapse_3}\footnote{Throughout this section and the discussion below, we use the term ``augmentation'' to refer to the integration of real and synthetic data during training}. In this approach, the $t$-th generative model is trained on a combination of a fixed real dataset $\mathcal{D}_0$ and the synthetic dataset $\widetilde{D}_{t-1}$ produced by the $(t-1)$-th generative model. \citet{bertrand2024on} show that this framework (Figure \ref{fig:ModelCollapse_3}) can effectively avoid the model collapse phenomenon if the proportion of synthetic data is small. In this paper, we refer to this framework (illustrated in Figure~\ref{fig:ModelCollapse_3}) as the \textit{synthetic augmentation} framework.

\begin{figure}[h]
    \centering
    \includegraphics[scale=0.26]{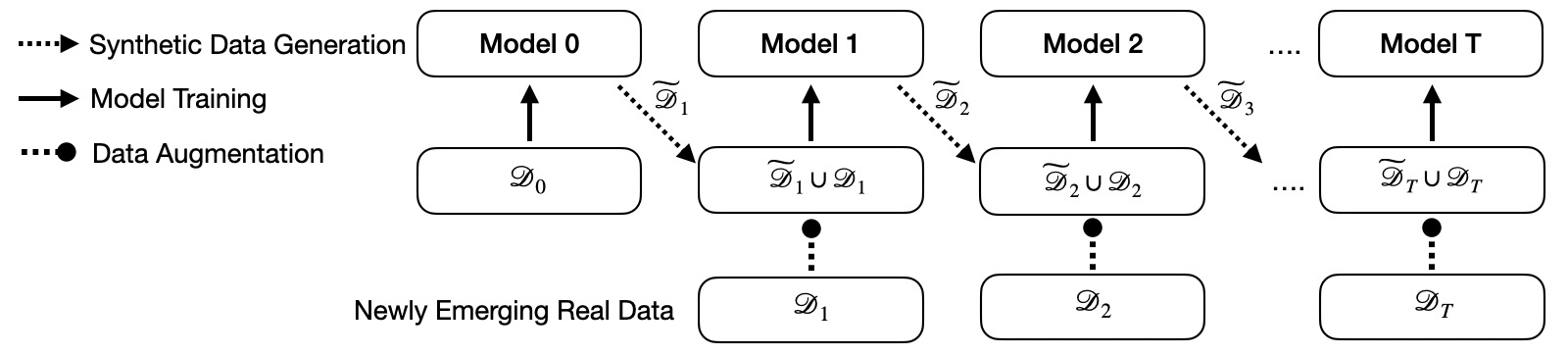}
    \caption{The general framework considered in this paper involves mixing newly collected real and synthetic data to address model collapse. In this framework, at the $t$-th training step, a newly collected dataset $\mathcal D_t$ is augmented with $\widetilde{\mathcal D}_{t}$ to train the $t$-th generative model.}
    \label{Fig:Framework}
\end{figure}

{
The frameworks illustrated in Figures \ref{fig:ModelCollapse_1}–\ref{fig:ModelCollapse_3} underscore a discouraging outcome: synthetic data do not enhance the performance of the final generative model, a finding also supported by our empirical results. The underlying reason is that, across these three frameworks, all synthetic samples ultimately derive from a fixed batch of real data. Consequently, the information they contain can never exceed that of the original $n$ real samples. This limitation is analogous to bootstrapping, which, while useful for variance estimation, cannot create information beyond what is contained in the original dataset.} Recently, \citet{alemohammad2024selfconsuming} investigate the phenomenon of model collapse in a \textit{new scenario}, where real data is continuously collected at each step for model estimation, as shown in Figure \ref{Fig:Framework}. In this work, the authors present empirical evidence demonstrating that a modest amount of synthetic data can enhance the performance of the final generative model. However, model performance declines when the amount of synthetic data exceeds a critical threshold. While \citet{alemohammad2024selfconsuming} provide empirical evidence for the effectiveness of synthetic data within the framework illustrated in Figure \ref{Fig:Framework}, the underlying theoretical foundation remains unexplored. In this paper, we refer to this framework (illustrated in Figure~\ref{Fig:Framework}) as the \textit{fresh data augmentation} framework.

In this paper, we theoretically investigate the phenomenon of model collapse within the new framework considered by \citet{alemohammad2024selfconsuming}. There are two primary practical reasons for studying the framework depicted in Figure \ref{Fig:Framework}. The first relates to privacy concerns. For instance, the California Consumer Privacy Act (CCPA)\footnote{\url{https://oag.ca.gov/privacy/ccpa}} grants users the right to request that platforms retain their personal information only for a limited time. Consequently, in practical applications, the real dataset $ \mathcal{D}_0$ may not always be available for future training as shown in the frameworks of Figures \ref{fig:ModelCollapse_1}-\ref{fig:ModelCollapse_3}. 
More precisely, our framework is most relevant when neither the raw historical data nor sufficiently informative reusable summaries of past data are available or compatible with future training.

Second, in a realistic setting, real data is continuously collected, enabling the training of new models using both synthetic data generated by prior models and newly collected real data. We refer to Section~A.1 in the supplementary file for additional discussion of practical scenarios in which
this framework is applicable.
  Despite its importance, the practice of training a model on a combination of newly collected real data and synthetic data from previous models remains theoretically underexplored. This naturally raises several critical questions:
\begin{itemize}[leftmargin=1cm] 
\item[{Q1}]: Can mixing newly collected real data with synthetic data during training help prevent model collapse?
\item[{Q2}]: What is the optimal ratio of real to synthetic data when training a new generative model with data from the previous model?
\item[{Q3}]: To what extent can synthetic data enhance estimation efficiency?
\end{itemize}

To address these questions, we first introduce our definition of model collapse in parametric generative modeling (see Definition~\ref{Def:MC}):  statistical inconsistency of parameter estimation caused by the recursive training process. To mitigate this issue, we propose a weighted training scheme for generative models that integrates newly collected real data with varying amounts of synthetic data.  This design is supported in practice by recent advances in watermarking techniques
\citep{cai2024towards,li2024robust,li2025statistical,li2025optimal,xie2025debiasing}
and probability-based detection methods \citep{mitchell2023detectgpt}; we refer to Section~A.3 in the supplementary file for further discussion.  We     investigate several scenarios of recursive parameter estimation. As a warm-up, we begin with Gaussian distribution estimation and then extend our analysis to generalized linear models, which encompass a broad class of learning problems as special cases. We also consider nonparametric distribution estimation. Methodologically, our analysis combines fine-grained finite-sample calculations—leveraging higher-order moment identities for simple distributions such as Gaussian with an asymptotic analysis for generalized linear models, derived from a weighted version of the asymptotically approximately linear expansion. In addition, we develop a nonparametric analysis under the Cram\'er–von Mises criterion.

% Our analysis demonstrates that the fresh data augmentation framework not only effectively prevents model collapse but also improves estimation efficiency. A comparison of different frameworks is provided in Table \ref{tab:simple_table}.
% \begin{table}[h!]
% \centering
% \caption{The utility of synthetic data in different frameworks illustrated in Figures \ref{fig:ModelCollapse_1}-\ref{Fig:Framework}}
% \begin{tabular}{l|c|c}
% \toprule
% {Framework} & {Model collapse?} & {Synthetic data enhance estimation?} \\ \hline
% Fully Synthetic     & \checkmark      & \ding{55}      \\ \hline
% Synthetic Accumulation      & \ding{55}     & \ding{55}   \\ \hline
% Synthetic Augmentation    & \ding{55}     & \ding{55}     \\ 
% \hline
% Fresh Data Augmentation    & \ding{55}     & \checkmark  \\
% \bottomrule
% \end{tabular}
% \label{tab:simple_table}
% \end{table}

{
In this paper, we summarize our contributions to the study of the model collapse phenomenon under the framework illustrated in Figure \ref{Fig:Framework} as follows:
 \begin{itemize}[leftmargin=0.7cm] 
     \item[(1)] {Unified Expression Across Models:} We show that across various scenarios, including Gaussian models, generalized linear models, and nonparametric estimation,the limiting error, under a given weight $w$ and data proportion, converges to a unified analytical expression. 
    \item[(2)] {Optimal Weight of Real and Synthetic Data:} Building on the unified expression of limit error, we derive the optimal weighting of real and synthetic data across regimes. We find that across Gaussian, GLM, and nonparametric settings, the optimal weight for a given data proportion admits a unified expression. Interestingly, naive unweighted mixing is always suboptimal and can be arbitrarily inefficient, especially when synthetic data dominates. Moreover, we identify threshold conditions on the weight of newly collected real data, above which model collapse is effectively prevented.
    \item[(3)] {Surprising Connection to the Golden Ratio:} When the amounts of real and synthetic data remain the same in each training iteration, we make a striking discovery: the optimal weight of real data is \textit{the reciprocal of the golden ratio}.
    \item[(4)] {Threshold for Beneficial Synthetic Data:} We demonstrate that as long as the weight of synthetic data in each iteration remains below a certain threshold, incorporating synthetic data enhances estimation efficiency in our fresh data augmentation setting compared to the MLE based only on the current real data batch, highlighting the effectiveness of incorporating synthetic data for training.
\end{itemize}
The remainder of this paper is organized as follows. Section \ref{Sec:Back_Pre} introduces the necessary notation and background on model collapse. Section \ref{Sec:GLM} presents the weighted recursive training framework, with a detailed analysis of Gaussian estimation and its extension to generalized linear models. Section \ref{Sec:EDE} investigates recursive training in a nonparametric distribution estimation setting. Section \ref{Sec:Exp} reports extensive experiments that validate our theoretical results. Proofs of the main theorems are provided in the supplementary file, which also contains additional results, discussions, and a comparison of our theoretical contributions with existing works.
}

\section{Preliminaries}
\label{Sec:Back_Pre}

In this section, we first introduce the necessary notations in Section \ref{Sec:Not}. To formalize our analysis, we then provide a mathematical definition of model collapse within the framework of parametric model estimation in Section \ref{Sec:MCD}. While model collapse was originally observed in the training of AI models \citep{shumailov2024ai}, a formal mathematical definition within the broader context of estimating parametric generative models remains absent.
\subsection{Notations}
\label{Sec:Not}
In this paper, we use bold notation to represent multivariate quantities and unbolded notation for univariate quantities. For any positive integer \( M \), we define \( [M] \) as the set $[M] = \{1,2,\ldots,M\}$. For example, $\bm{x} \in \mathbb{R}^p$ represents a $p$-dimensional vector, while $x$ represents a real value. For a vector $\bm{x} \in \mathbb{R}^p$, we denote its %$l_1$-norm and 
$l_2$-norm as  
%$\Vert \bm{x} \Vert_1=\sum_{i=1}^p |x_i|$ and 
$\Vert \bm{x} \Vert_2=\big(\sum_{i=1}^p |x_i|^2\big)^{1/2}$. 
%For two given sequences $\{A_n\}_{n \in \mathbb{N}}$ and $\{ B_n\}_{n\in \mathbb{N}}$, we write $A_n \gtrsim B_n$ if there exists a constant $C>0$ such that $A_n \geq C B_n$ for all $n \in \mathbb{N}$. 
%Additionally, we write $A_n \asymp B_n$ if both $A_n \gtrsim B_n$ and $A_n \lesssim B_n$ hold. 
For a continuous random variable $X$, we let $P_X(x)$ denote its probability density function at $x$ and $\mathbb{P}_{X}$ denote the associated probability measure. We use $\mathbb{E}_{X}$ to denote the expectation taken with respect to the randomness of $X$. For a sequence of random variables $\{X_n\}_{n\in \mathbb{N}}$, we write $X_n =o_p(1)$ to indicate that $X_n$ converges to zero in probability. For a matrix $\bm{A}=(A_{ij})_{i,j \in [n]}$, we define its trace as $\text{tr}(\bm{A})=\sum_{i=1}^n A_{ii}$. Finally, we use $\mathcal{F}_t$ to represent the $\sigma$-algebra generated by all events occurring up to the $t$-th training step.

\subsection{Model Collapse in Parametric Generative Models}
\label{Sec:MCD}
\begin{figure}[h]
    \centering
    \includegraphics[scale=0.25]{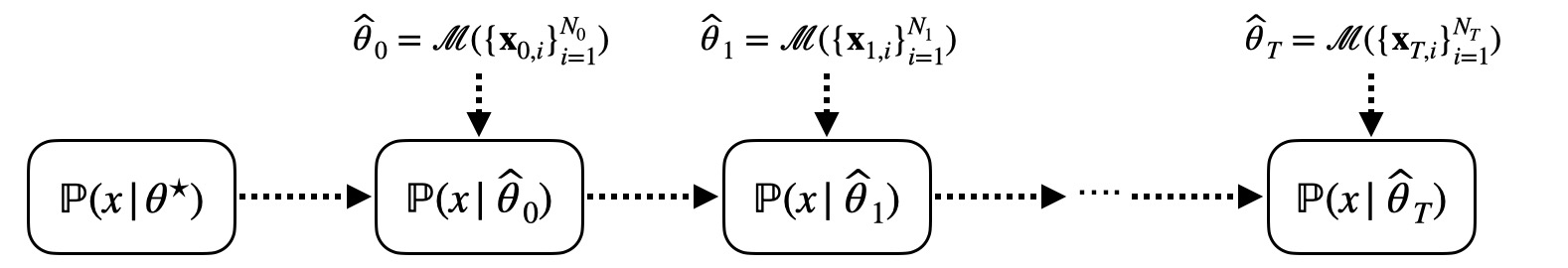}
    \caption{The general framework of recursive model estimation.}
    \label{fig:UF}
\end{figure}

In this section, we present a general definition of \textit{model collapse} for parametric generative models within the framework of recursive model estimation as illustrated in Figure \ref{fig:UF}. Consider a family of generative models parameterized by $\bm{\theta} \in \bm{\Theta}$, denoted as $\{\mathbb{P}(\cdot \mid \bm{\theta}) \mid \bm{\theta} \in \bm{\Theta}\}$. Let $\bm{\theta}^\star$ be the ground truth parameter. The goal is to estimate $\bm{\theta}^\star$ after successive iterations.

In the framework of Figure \ref{fig:UF}, an estimation scheme $\mathcal{M}: \mathcal{X}^{\mathbb{N}} \to \bm{\Theta}$ is applied consistently at each estimation step, i.e., $\widehat{\bm{\theta}}_t = \mathcal{M}(\{\bm{x}_{t,i}\}_{i=1}^{N_t})$ for $t \geq 0$, where $\mathcal{X}$ represents the data support, $\bm{\Theta}$ denotes the parameter space, and $\{\bm{x}_{t,i}\}_{i=1}^{N_t}$ is the dataset used at step $t$. Aligning with the frameworks in Figure \ref{fig:ModelCollapse_1}-\ref{Fig:Framework}, we define $\{\bm{x}_{0,i}\}_{i=1}^{N_0}$ as a real dataset drawn from $\mathbb{P}(\bm{x} \mid \bm{\theta}^\star)$. For $t \geq 1$, $D_t$ may consist of either real data or synthetic data generated by previous generative models $\{\mathbb{P}(\bm{x} \mid \widehat{\bm{\theta}}_{i}) : 0 \leq i \leq t-1\}$. Without loss of generality, we assume that the sample sizes $\{N_t\}_{t=0}^{\infty}$ follow the pattern $N_t = N_0 \times c_t$ for $t \geq 1$. In other words, given a fixed initial real sample size $N_0$, the sequence $\{c_t\}_{t=1}^{\infty}$ determines the trend of the training datasets throughout the training process. 
In this work, we focus on the setting where $c_t \equiv c$ is a constant independent of $t$, so that the dataset size remains the same at each round.

We denote by $\mathbb{P}_{t}$ the underlying distribution of $\{\bm{x}_{t,i}\}_{i=1}^{N_t}$.  Since $\{\bm{x}_{t,i}\}_{i=1}^{N_t}$ consists of both real samples, which may include newly
collected real data and historical real data, and synthetic samples generated by
previous models, 
 $\mathbb{P}_{t}$ is induced by combining the real
data distribution $\mathbb{P}(\bm{x}\mid\bm{\theta}^\star)$ and data generated from the
sequence of past models
$\{\mathbb{P}(\bm{x}\mid\widehat{\bm{\theta}}_{i}) : 0 \le i \le t-1\}$. In other words, the dependence of $\mathbb{P}_{t}$ on $\mathbb{P}(\bm{x} \mid \bm{\theta}^\star)$ and $\{\mathbb{P}(\bm{x} \mid \widehat{\bm{\theta}}_{i}) : 0 \leq i \leq t-1\}$ is determined by the strategy used to integrate synthetic and real data for training. 

The framework in Figure \ref{fig:UF} generalizes those in Figures \ref{fig:ModelCollapse_1}-\ref{Fig:Framework} as special cases. Particularly, the framework in Figure \ref{fig:UF} simplifies to the \textit{fully-synthetic case} in Figure \ref{fig:ModelCollapse_1} when $N_i = N_j$ for all $i, j \geq 0$, and $\{\bm{x}_{t,i}\}_{i=1}^{N_t}$ is entirely sampled from $\mathbb{P}(\bm{x} \mid\widehat{\bm{\theta}}_{t-1})$, i.e., $\mathbb{P}_{t}(\bm{x})=\mathbb{P}(\bm{x} \mid \widehat{\bm{\theta}}_{t-1})$ for $t \geq 1$. The key challenge in Figure \ref{fig:UF} is to assess the performance of the final generative model $\mathbb{P}(\cdot \mid \widehat{\bm{\theta}}_T)$ when $T$ is large, which can be quantified by the estimation error $\mathbb{E} \|\widehat{\bm{\theta}}_T - \bm{\theta}^\star\|_2^2$. Here, the expectation is taken with respect to the randomness of all previous training datasets. If $N_0$ is fixed, the performance of $\widehat{\bm{\theta}}_T$ in estimating $\bm{\theta}^\star$ as $T \to \infty$ is influenced by three key factors: (1) the estimation scheme $\mathcal{M}(\cdot)$, (2) the pattern of sample sizes $\{c_t\}_{t=1}^\infty$, and (3) the underlying distribution $\mathbb{P}_t$ at each step. In other words, once $(\mathcal{M}, \{c_t\}_{t=1}^{\infty}, \{\mathbb{P}_t\}_{t=1}^{\infty})$ is specified, the training strategy for recursive estimation is uniquely determined.

In statistical learning theory, if $(\mathcal{M}, \{c_t\}_{t=1}^{\infty}, \{\mathbb{P}_t\}_{t=1}^{\infty})$ is fixed with an appropriate estimation scheme $\mathcal{M}$, the error $\mathbb{E} \|\widehat{\bm{\theta}}_T - \bm{\theta}^\star\|_2^2$ for a fixed $T$ should converge to zero as $N_0$ increases to infinity. However, existing literature \citep{shumailov2024ai,kazdan2024collapse} shows that, for any fixed $N_0$, as $T$ increases, $\widehat{\bm \theta}_T$ gradually loses information about $\mathbb{P}(\cdot|\bm{\theta}^\star)$ and finally fails to converge to $\bm{\theta}^\star$ regardless of the size of data during recursive training process. Therefore, preventing model collapse should focus on designing appropriate data integration and training schemes to ensure that the estimation error \( \mathbb{E} \|\widehat{\bm{\theta}}_T - \bm{\theta}^\star\|_2^2 \) remains stable as \( T \) grows to infinity for any fixed $N_0$, and decreases as more data is used for training. Based on this rationale, we formally define model collapse in the context of parametric generative models in Definition \ref{Def:MC}.

\begin{definition}[Model Collapse in Parametric Generative Models] 
\label{Def:MC}
For a recursive training scheme $(\mathcal{M}, \{c_t\}_{t=1}^{\infty}, \{\mathbb{P}_t\}_{t=1}^{\infty})$, we say that model collapse occurs if there exists a constant \(\delta > 0\) such that $\limsup_{T \to \infty}
\mathbb{E} \left[
\big\|\widehat{\bm{\theta}}_T - \bm{\theta}^\star\big\|_2^2
\right]
\ge \delta$
 for any value of $N_0$.
\end{definition}

In Definition \ref{Def:MC}, model collapse refers to the phenomenon where the estimator \(\widehat{\bm{\theta}}\) progressively loses information about the true data distribution \(\mathbb{P}(\cdot|\bm{\theta}^\star)\) as \(T\) increases, causing \(\widehat{\bm{\theta}}\) to fail to converge to \(\bm{\theta}^\star\) regardless of the samples used throughout the training process. We characterize this failure by the estimator’s inconsistency, independent of the sample sizes used at each estimation step.  Specifically, the limiting error $\limsup_{T \to \infty}
\mathbb{E} \left[
\big\|\widehat{\bm{\theta}}_T - \bm{\theta}^\star\big\|_2^2
\right]$
remains bounded away from zero, no matter how large the sample sizes are at any step.

{
\section{Weighted Recursive Training}
\label{Sec:GLM}
In this section, we mainly analyze the following two scenarios under the proposed fresh data augmentation framework. Specifically, we begin with a \textit{warm-up example} based on the Gaussian estimation problem. The Gaussian setting serves as a natural starting point due to its analytical tractability, as the estimator admits a closed-form solution. We then extend our findings to a broader class of \textit{generalized linear models} (GLMs), where analogous recursive training dynamics in the asymptotic regime can be rigorously characterized. Through these examples, we uncover the following insights for recursive training:
\begin{itemize}[leftmargin=1.2cm, itemindent=0.8cm]
    \item[{Insight 1}:] In recursive training regimes where synthetic data are repeatedly generated and incorporated into subsequent training, weighted training scheme leads to substantially improved performance compared to naive joint training.
    \item[{Insight 2}:] When the size of synthetic data generated at each recursion equals that of the real data, the optimal weighting of real data in the final estimator exhibits a surprising connection to the golden ratio.
\end{itemize}
}

\subsection{Multivariate Gaussian}
\label{SubSec:Gaus2}
In this section, we investigate the phenomenon of model collapse within the fresh data augmentation framework, using recursive Gaussian distribution estimation as a warm-up. Specifically, we assume the real data distribution follows \( \mathcal{N}(\bm{\mu}, \bm{\Sigma}) \), where \( \bm{\mu} \in \mathbb{R}^p \) is the mean vector and \( \bm{\Sigma} \in \mathbb{R}^{p \times p} \) is the covariance matrix. The goal of recursive estimation is to estimate \( \bm{\mu} \) and \( \bm{\Sigma} \) simultaneously. The overall procedure of this case is summarized in {Fresh Data Augmentation Case}~\ref{alg:Multivariable_gaussian_estimation}.

{Estimation.} Let \( \mathcal{D}_{t} = \{\bm{x}_{t,i}\}_{i=1}^n \) denote the real dataset collected at the \( t \)-th training step, and let \( \widetilde{\mathcal{D}}_{t} = \{\widetilde{\bm{x}}_{t,i}\}_{i=1}^m \) represent the synthetic dataset generated during the \((t-1)\)-th generation step. This setup is designed to mimic a realistic scenario in which, at each stage of training a new generative model, both synthetic data and newly collected real data are available for training. To estimate the underlying Gaussian distribution, we adopt the following weighted estimation scheme at the \( t \)-th ($t\geq 1$)  training step:
\begin{align}
     \text{Mean Estimation: } &   \boldsymbol{\mu}_{t}(w,n,m) = \argmin_{\boldsymbol{\mu} \in \mathbb{R}^p}  \frac{w}{n} \sum_{i=1}^n \| \boldsymbol{\mu} - \boldsymbol{x}_{t,i} \|_2^2 + \frac{1-w}{m} \sum_{i=1}^m \| \boldsymbol{\mu} - \widetilde{\boldsymbol{x}}_{t,i} \|_2^2 , \label{MeanEsti}\\
     \text{Variance Estimation: } &  \boldsymbol{\Sigma}_t(w,n,m) = \argmin_{\boldsymbol{\Sigma} \in \mathbb{R}^{p \times p}}  w \| \boldsymbol{\Sigma} - \widehat{\boldsymbol{S}}_t \|_F^2 + (1-w) \| \boldsymbol{\Sigma} - \widetilde{\boldsymbol{S}}_t \|_F^2 ,
\end{align}
where $k = \frac{n}{m}$ denotes the ratio of real to synthetic sample sizes, \( \widehat{\boldsymbol{S}}_t =  \frac{1}{n-1}\sum_{i=1}^n (\boldsymbol{x}_{t,i}-\widehat{\boldsymbol{\mu}}_t)(\boldsymbol{x}_{t,i}-\widehat{\boldsymbol{\mu}}_t)^\top \) and \( \widetilde{\boldsymbol{S}}_t =  \frac{1}{m-1}\sum_{i=1}^m (\widetilde{\boldsymbol{x}}_{t,i}-\widetilde{\boldsymbol{\mu}}_t)(\widetilde{\boldsymbol{x}}_{t,i}-\widetilde{\boldsymbol{\mu}}_t)^\top \) with \( \widehat{\boldsymbol{\mu}}_t=\frac{1}{n}\sum_{i=1}^n \boldsymbol{x}_{t,i} \) and \( \widetilde{\boldsymbol{\mu}}_{t}=\frac{1}{m}\sum_{i=1}^m \widetilde{\boldsymbol{x}}_{t,i} \). At the initial step $t=0$, since no synthetic samples are available, the weight on real data is effectively $w=1$. Consequently, the estimator reduces to the standard maximum likelihood estimator based on the initial real dataset, which serves as the base case for the recursive weighted procedure.

{Evaluation.} We define the following metrics to analyze the quantitative behavior of $\bm{\mu}_t(w,n,m)$ and $\bm{\Sigma}_t(w,n,m)$ especially when $t$ increases to infinity:
\begin{align*}
    \text{Err}(\bm{\mu}_t(w,n,m)) = \mathbb{E}\big[\Vert \bm{\mu}_t(w,n,m) - \bm{\mu}\Vert_2^2\big] \mbox{ and }
    \text{Err}(\bm{\Sigma}_t(w,n,m)) = \mathbb{E}\big[\Vert \bm{\Sigma}_t(w,n,m) - \bm{\Sigma}\Vert_F^2\big],
\end{align*}  
where $\Vert \cdot \Vert_2$ is the $l_2$-norm and $\Vert \cdot \Vert_F$ is the Frobenius norm. Here, $\text{Err}(\bm{\mu}_{t}(w,n,m))$ and $\text{Err}(\bm{\Sigma}_t(w,n,m))$ quantify the estimation errors of $\bm{\mu}_{t}(w,n,m)$ and $\bm{\Sigma}_t(w,n,m)$, respectively. The expectation is taken with respect to the randomness inherent in all real and synthetic datasets generated during the first $t$ steps. The central problem is analyzing the behavior of $\bm{\mu}_t(w,n,m)$ and $\bm{\Sigma}_t(w,n,m)$ as $t$ approaches infinity. Therefore, we also consider the following limiting metrics for the case where \( t = \infty \):
\begin{align*}
    \mathrm{Err}(\bm{\mu}_{\infty}(w,n,m)) \triangleq \lim_{t \rightarrow \infty} \mathrm{Err} \left( \bm{\mu}_{t}(w,n,m) \right) \mbox{ and } \mathrm{Err}(\bm{\Sigma}_{\infty}(w,n,m)) \triangleq \lim_{t \rightarrow \infty} \mathrm{Err} \left( \bm{\Sigma}_t(w,n,m) \right).
\end{align*}

\begin{algorithm}[h]
\caption{- Multivariate Gaussian Estimation}
\label{alg:Multivariable_gaussian_estimation}
\begin{algorithmic}
\STATE {Initialization:} Compute \( \boldsymbol{\mu}_0 = \frac{1}{n}\sum_{i=1}^n \boldsymbol{x}_{0,i} \) and \( \boldsymbol{\Sigma}_0 = \frac{1}{n-1}\sum_{i=1}^n (\boldsymbol{x}_{0,i}-\boldsymbol{\mu}_0)(\boldsymbol{x}_{0,i}-\boldsymbol{\mu}_0)^T \) based on \( \mathcal{D}_0=\{\boldsymbol{x}_{0,i}\}_{i=1}^n \) from \( N(\boldsymbol{\mu},\boldsymbol{\Sigma}) \);
\FOR{$t = 1,2,\ldots,T$}
    \STATE {1. Data Collection}: Sample \( \widetilde{\mathcal{D}}_{t}=\{\widetilde{\boldsymbol{x}}_{t,i}\}_{i=1}^m \) from \( N(\boldsymbol{\mu}_{t-1},\boldsymbol{\Sigma}_{t-1}) \) and collect a new real dataset \( \mathcal{D}_{t}=\{\boldsymbol{x}_{t,i}\}_{i=1}^n \) from \( N(\boldsymbol{\mu},\boldsymbol{\Sigma}) \);
    \STATE {2. Mean Estimation}: Compute the mean of the Gaussian distribution at the $t$-th training step:
    \begin{equation}
        \label{Eqn:MULMeanUpdate}
         \boldsymbol{\mu}_{t}(w,n,m) =  w \widehat{\boldsymbol{\mu}}_t + (1-w)\widetilde{\boldsymbol{\mu}}_{t},
    \end{equation}
   where \( \widehat{\boldsymbol{\mu}}_t=\frac{1}{n}\sum_{i=1}^n \boldsymbol{x}_{t,i} \), \( \widetilde{\boldsymbol{\mu}}_{t}=\frac{1}{m}\sum_{i=1}^m \widetilde{\boldsymbol{x}}_{t,i} \), and \( k = n/m \).

    \STATE {3. Covariance Estimation}: Compute the covariance of the Gaussian distribution at the $t$-th training step:
        \begin{equation}
        \label{Eqn:MULVarUpdate}
\boldsymbol{\Sigma}_t(w,n,m) = w\widehat{\boldsymbol{S}}_t + (1-w)\widetilde{\boldsymbol{S}}_t.,
    \end{equation}
    where \( \widehat{\boldsymbol{S}}_t =  \frac{1}{n-1}\sum_{i=1}^n (\boldsymbol{x}_{t,i}-\widehat{\boldsymbol{\mu}}_t)(\boldsymbol{x}_{t,i}-\widehat{\boldsymbol{\mu}}_t)^T \) and \( \widetilde{\boldsymbol{S}}_t =  \frac{1}{m-1}\sum_{i=1}^m (\widetilde{\boldsymbol{x}}_{t,i}-\widetilde{\boldsymbol{\mu}}_t)(\widetilde{\boldsymbol{x}}_{t,i}-\widetilde{\boldsymbol{\mu}}_t)^T \).
\ENDFOR
\RETURN $\boldsymbol{\mu}_T(w,n,m)$ and $\boldsymbol{\Sigma}_T(w,n,m)$.
\end{algorithmic}
\end{algorithm}
The primary motivation for examining the limiting errors stems from the continual evolution of languages models, which naturally raises the question of what future language models might look like \citep{shumailov2024ai}. Furthermore, prior studies have identified \emph{model collapse} as a phenomenon in which estimation errors explode in parametric settings as the recursive training process unfolds \citep{dey2024universality,dohmatob2024model}. Hence, we are interested in how estimation errors evolve throughout the recursive training process.

{Theoretical Analysis.} In the following, we present the theoretical results (Theorems \ref{Thm1:Mul}) on the quantitative behavior of $\mathrm{Err}(\bm{\mu}_{\infty}(w,n,m))$ and $\mathrm{Err}(\bm{\Sigma}_{\infty}(w,n,m))$ in a finite-sample setting, aiming to elucidate the impact of $w$ and $k$ on the estimation of Gaussian distribution during recursive training. Particularly, we find that $\mathrm{Err}(\bm{\Sigma}_{\infty}(w,n,m))$ can be succinctly characterized by two key quantities: \(\mathrm{tr}(\mathbf{\Sigma}^2)\) and \(\left[\mathrm{tr}(\mathbf{\Sigma})\right]^2\).

\begin{theorem}
    \label{Thm1:Mul}
For any $w \in (0,1]$ and $n,m \geq 2$ with $k=\frac{n}{m}$.
It holds that 
\begin{align*}
\mathrm{Err}(\bm{\mu}_{\infty}(w,n,m)) &= \lim_{t \rightarrow \infty} \mathrm{Err}(\bm{\mu}_{t}(w,n,m) ) 
=\frac{C(w,k)\mathrm{tr}(\bm{\Sigma})}{n}, \\
      \mathrm{Err}(\bm{\Sigma}_{\infty}(w,n,m))&=
      \lim_{t \rightarrow \infty} \mathrm{Err}(\bm{\Sigma}_{t}(w,n,m) ) 
      =
  \begin{cases} 
\infty, & 0 < w \leq 1 - \sqrt{\frac{m-1}{m+1}}, \\
\frac{N}{D}-\mathrm{tr}(\bm{\Sigma}^2), & 1 - \sqrt{\frac{m-1}{m+1}} < w \leq 1,
\end{cases}
\end{align*}
where $C(w,k) =\frac{w^2+(1-w)^2k}{2w-w^2}$, and \( N \) and \( D \) are defined as
\begin{align*}
&N  = (2w-w^2)\mathrm{tr}(\bm{\Sigma}^2)+
\underbrace{\frac{(w-1)^2\left[
\mathrm{tr}^2(\bm{\Sigma})(2w-w^2)
+ \frac{2w^2\mathrm{tr}(\bm{\Sigma}^2)}{n-1}
\right] }{(m-1)(2w-w^2)}}_{\text{Convergence to 0 as } m \rightarrow \infty}
+ \frac{w^2[\mathrm{tr}^2(\bm{\Sigma})+\mathrm{tr}(\bm{\Sigma}^2)]}{n-1},\\
&D = 
2w-w^2
- \underbrace{\frac{2(w-1)^4}{(m-1)^2[2w-w^2]} - \frac{(w-1)^2}{m-1}}_{\text{Convergence to 0 as } m \rightarrow \infty}.
\end{align*}
\end{theorem}

In Theorem~\ref{Thm1:Mul}, we derive the exact expressions for both the mean and covariance estimation errors in the finite-sample setting, $\mathrm{Err}(\bm{\mu}_{\infty}(w,n,m))$ and $\mathrm{Err}(\bm{\Sigma}_{\infty}(w,n,m))$, under different values of the weighting parameter $w$, the size of real data $n$ and the size of synthetic data $m$ in each round. For the mean estimator, it is interesting to note that as long as $w > 0$ and $k > 0$, the estimation error remains finite and converges to $0$ as $n \rightarrow \infty$. This result indicates that model collapse is prevented due to the incorporation of new real data. For the covariance estimator in the finite-sample setting, $\mathrm{Err}(\bm{\Sigma}_{\infty}(w,n,m))$ exhibits the following two regimes:
\begin{itemize}
    \item[(1)] When $w \le 1 - \sqrt{\tfrac{m-1}{m+1}}$ (assigning insufficient weight to real samples), the estimation error $\mathrm{Err}(\bm{\Sigma}_{\infty}(w,n,m))$ diverges to infinity, leading to \emph{model collapse}. 
    \item[(2)] When $w > 1 - \sqrt{\tfrac{m-1}{m+1}}$, 
    $\mathrm{Err}(\bm{\Sigma}_{\infty}(w,n,m))$ remains finite for any $(n,m)$, 
    indicating that the proposed weighting scheme effectively mitigates model collapse in covariance estimation.
\end{itemize}

Even though Theorem \ref{Thm1:Mul} shows that incorporating newly collected real data can prevent exploding estimation errors under an appropriate choice of $w$, two natural questions arise from these results:
\begin{itemize}[leftmargin=2cm, itemindent=0.8cm]
    \item[{Question 1}:] For a fixed ratio  scheme (i.e., fixed $k$), what are the optimal weights that minimize $\mathrm{Err}(\bm{\mu}_{\infty}(w,n,m))$ and $\mathrm{Err}(\bm{\Sigma}_{\infty}(w,n,m))$, respectively?
    \item[{Question 2}:] Compared with using only new real data, can incorporating synthetic data improve the estimation accuracy?
\end{itemize}
To answer the above questions, we present the following theoretical results (Corollary \ref{Coro:Multi_Mean}).

\begin{corollary}
    \label{Coro:Multi_Mean}
    For any $k>0$, $\mathrm{Err}(\bm \mu_{\infty}(w,n,m))$ and $\mathrm{Err}(\bm \Sigma_{\infty}(w,n,m))$ exhibit the following cases:
        \begin{itemize}
        \item[(1)] \textnormal{{Fully Synthetic - Model Collapse}}:  
        \begin{align*}
            \lim_{w \rightarrow 0^{+}} \mathrm{Err}(\bm{\mu}_{\infty}(w,n,m)) = \infty \,\mbox{ and }\,
            \lim_{w \rightarrow 0^{+}} \mathrm{Err}(\bm{\Sigma}_{\infty}(w,n,m)) = \infty.
        \end{align*}
        \item[(2)] \textnormal{{Using Only One Round of Real Data}}: 
        \begin{align*}
            \mathrm{Err}(\bm{\mu}_{\infty}(1,n,m)) = \frac{\mathrm{tr}(\bm{\Sigma})}{n} \,\mbox{ and }\,
            \mathrm{Err}(\bm{\Sigma}_{\infty}(1,n,m)) = \frac{\mathrm{tr}(\bm{\Sigma}^2)+\mathrm{tr}^2(\bm{\Sigma})}{n-1}.
        \end{align*}
        \item[(3)] \textnormal{{Optimal Weight for Mean Estimation}}: 
        \begin{align*}
            w^\star = \argmin_{w \in [0,1]} \,\mathrm{Err}(\bm{\mu}_{\infty}(w,n,m)) =  \frac{\sqrt{k^{2} + 4k} - k}{2} \Longrightarrow
            \mathrm{Err}(\bm{\mu}_{\infty}(w^\star,n,m)) =
            \frac{w^\star\mathrm{tr}(\bm{\Sigma})}{n}
        \end{align*}
        \item[(4)] \textnormal{{Asymptotic Optimal Weight for Variance Estimation}}: Given $n,m \to \infty$ with $\frac{n}{m} \to k$, we have
\begin{align*}
    \mathrm{Err}(\bm{\Sigma}_{\infty}(w,n,m)) 
    = \frac{ C(w,k) \left[\mathrm{tr}(\bm{\Sigma}^2) + \mathrm{tr}^2(\bm{\Sigma})\right] }{n} + o(n^{-1}).
\end{align*}
This result implies that, in the asymptotic regime, $\mathrm{Err}(\bm{\Sigma}_{\infty}(w,n,m))$ is minimized at $w^\star = \frac{\sqrt{k^{2} + 4k} - k}{2}$ for any $k > 0$.
\item[(5)] \textnormal{{Estimation Improvement}}: Using $w=w^\star$, for any $k \in \mathbb{R}_{>0}$, we have 
\begin{align*}
\frac{\mathrm{Err}(\bm{\mu}_{\infty}(w^\star,n,m))}{\mathrm{Err}(\bm{\mu}_{\infty}(1,n,m))}=
\lim_{m,n\rightarrow \infty,\frac{n}{m}\rightarrow k}
    \frac{\mathrm{Err}(\bm{\Sigma}_{\infty}(w^\star,n,m))}{\mathrm{Err}(\bm{\Sigma}_{\infty}(1,n,m))} = \frac{\sqrt{k^2+4k}-k}{2}<1.
\end{align*}
    \end{itemize}
\end{corollary}

Several important implications follow from Corollary \ref{Coro:Multi_Mean}.  
First, when $w = 0$ (the fully synthetic case), both $\mathrm{Err}(\bm{\mu}_{\infty}(w,n,m))$ and $\mathrm{Err}(\bm{\Sigma}_{\infty}(w,n,m))$ diverge to infinity, indicating that relying solely on synthetic data inevitably leads to model collapse \citep{dey2024universality,shumailov2024ai}. Second, for mean estimation, Corollary \ref{Coro:Multi_Mean} shows that the optimal weight is $w^\star = \frac{\sqrt{k^2 + 4k} - k}{2}$, which yields a strictly smaller estimation error than using  a single round of real data while discarding all others. This improvement is guaranteed by the inequality 
$\frac{\sqrt{k^2 + 4k} - k}{2} < 1$ for any $k > 0$. Third, for variance estimation, it is challenging to derive the optimal weight in closed form in the finite-sample regime. To address this, we analyze the asymptotic behavior of $\mathrm{Err}(\bm{\Sigma}_{\infty}(w,n,m))$ as $m,n \to \infty$ while maintaining $\frac{n}{m} \to k$. In this regime, the error is dominated by 
$\frac{C(w,k)\left[\mathrm{tr}(\bm{\Sigma}^2) + \mathrm{tr}^2(\bm{\Sigma})\right]}{n}$,  
and the minimizing weight coincides with the same $w^\star$ as above.

Here, the optimality of \(w^\star\) is restricted to the class of constant-in-time weighting schemes, where the same weight \(w\) is used at every recursive training step. When adaptive or horizon-dependent weights are allowed, the recursive dynamics may exhibit different behavior, as discussed in Remark~\ref{remark_depend_T} and Section~A.7.

\begin{figure}[h!]
    \centering
    \begin{subfigure}[b]{0.485\textwidth}
        \centering
        \includegraphics[width=\textwidth]{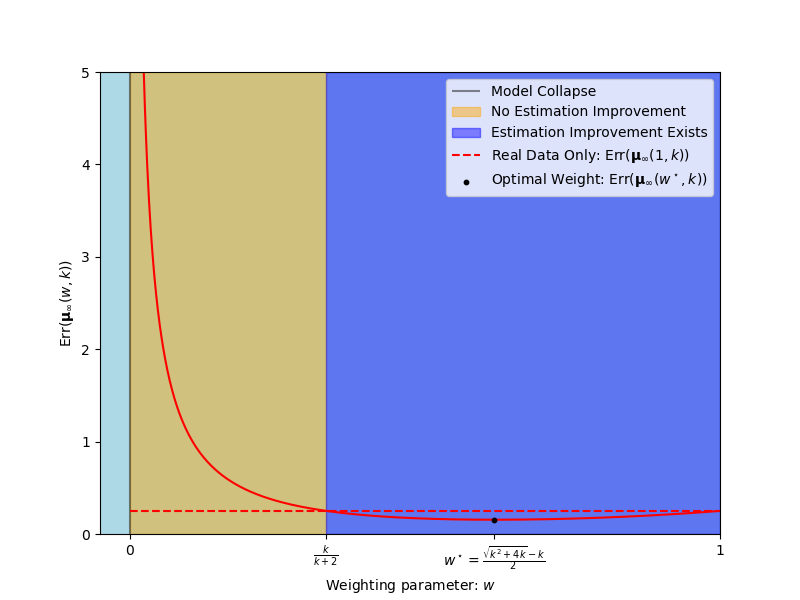}
        \caption{Crossover Points of $\mathrm{Err}(\bm{\mu}_{\infty}(w,n,m))$}
    \end{subfigure}
        \hfill
    \begin{subfigure}[b]{0.485\textwidth}
        \centering
        \includegraphics[width=\textwidth]{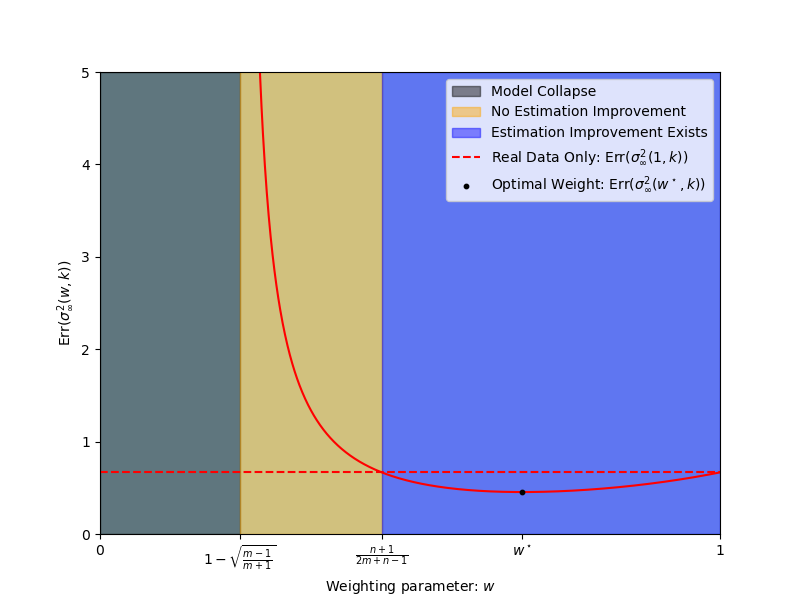}
        \caption{Crossover Points of $\mathrm{Err}(\bm{\Sigma}_{\infty}(w,n,m))$}
    \end{subfigure}
    \caption{The figure illustrates the quantitative behavior of $\mathrm{Err}(\bm{\mu}_{\infty}(w,n,m))$ and $\mathrm{Err}(\bm{\Sigma}_{\infty}(w,n,m))$ with $p=1$ across three stages: (1) the model collapse stage, (2) the stage with no estimation improvement, and (3) the stage with estimation improvement.}
    \label{fig:Sigma}
\end{figure}

Finally, Corollary \ref{Coro:Multi_Mean} establishes that using $w=w^\star$ leads to consistent improvement in both mean and variance estimation. In particular, for any $k>0$, we confirm that incorporating synthetic data with an optimally chosen weight strictly improves estimation accuracy compared with using only newly collected real data. The impact of $w$ on estimation errors for a fixed $k$ is summarized in Figure \ref{fig:Sigma}.

{Optimality of Weighting Scheme.} A natural question arises as to whether the weighting scheme is optimal. To address this, we demonstrate that the proposed weighted combination in Theorem~\ref{Thm1:Mul} retains the property of being the Best Linear Unbiased Estimator, as formally stated in the following proposition.

\begin{proposition}
\label{corollary_global_optimum}
Consider a more general update rule for mean estimation given by:\begin{equation}
    \label{Eqn:MULMeanUpdate_General_multi}
    \bm{\mu}_{t}(\{w_i\}_{i=1}^n,\{\widetilde{w}_j\}_{j=1}^m,n,m) = \frac{1}{n}\sum_{i=1}^n w_i \bm{x}_{t,i}+\frac{1}{m}\sum_{j=1}^m \widetilde{w}_j\widetilde{\bm{x}}_{t,i},
\end{equation}
where \( w_i \) and \( \widetilde{w}_j \) are the weights assigned to the \( i \)-th real and the $j$-th synthetic data points, respectively, satisfying that $\sum_{i=1}^n w_i/n + \sum_{j=1}^m \widetilde{w}_j/m = 1$. Consider the limiting estimation error defined as
$$\mathrm{Err}(\bm{\mu}_{\infty}(\{w_i\}_{i=1}^n,\{\widetilde{w}_j\}_{j=1}^m,n,m)) = \lim_{t\rightarrow \infty}\mathbb{E}\big[\Vert \bm{\mu}_{t}(\{w_i\}_{i=1}^n,\{\widetilde{w}_j\}_{j=1}^m,n,m) - \bm{\mu}\Vert_2^2\big].$$ It holds true that $\mathrm{Err}(\bm{\mu}_{\infty}(\{w_i\}_{i=1}^n,\{\widetilde{w}_j\}_{j=1}^m,n,m))$ is minimized by $w_i = w^{\star}$ for $i \in [n]$ and $\widetilde{w}_j = 1 - w^{\star}$ for $j \in [m]$, where \( w^{\star}= \frac{\sqrt{k^{2} + 4k} - k}{2} \) as defined in Theorem \ref{Thm1:Mul}.

\end{proposition}

\noindent
\begin{remark}
Proposition \ref{corollary_global_optimum} establishes the optimality of $w^\star$ in recursive training among all possible weighted training schemes for estimating the Gaussian mean. An immediate implication of this result is that, during recursive training, maintaining the distinguishability between real and synthetic data is crucial for obtaining a more accurate estimator. In contrast, if we directly merge the datasets $\mathcal{D}_t$ and $\widetilde{\mathcal{D}}_t$ and minimize the empirical squared loss jointly:
\begin{align*}
\boldsymbol{\mu}_{t}(w,n,m) = \argmin_{\boldsymbol{\mu} \in \mathbb{R}^p}  \frac{1}{n+m} \left\{ \sum_{i=1}^n \| \boldsymbol{\mu} - \boldsymbol{x}_{t,i} \|_2^2 +\sum_{i=1}^m \| \boldsymbol{\mu} - \widetilde{\boldsymbol{x}}_{t,i} \|_2^2 \right\},
\end{align*}
then the resulting estimate can be written as $\bm{\mu}_{t}(w_0, n,m) =  w_0 \widehat{\bm{\mu}}_t + (1-w_0)\widetilde{\bm{\mu}}_{t}$, where the implicit weight \(w_0\) is given by:
\begin{equation*}
w_0 = \frac{n}{n+m}  = \frac{k}{k+1} 
 \neq w^{\star} = \frac{\sqrt{k^2+4k}-k}{2}, \text{ for any } k \in (0,+\infty)
\end{equation*}
A detailed derivation of this result is provided in the supplementary file. The inequality implies that, in multivariate Gaussian estimation, directly mixing the data yields a strictly suboptimal weight for mean estimation for any value of $k$.
\end{remark}

\begin{remark}
In the Gaussian mean estimation setting, incorporating synthetic data through weighted training admits an explicit interpretation as an $\ell_2$-type regularization toward an anchor induced by the previous model. We refer to Section~A.6 in supplementary file for a detailed derivation.
\end{remark}

\begin{remark}
\label{remark_depend_T}
An additional implication of Corollary~\ref{Coro:Multi_Mean} is that, for any fixed weight
$w \in (0,1]$, and fixed real and synthetic sample sizes $n$ and $m$, letting $T \to \infty$ leads to a non-vanishing limiting error.
This error floor arises from the imposed data-retention constraint in the fresh data
augmentation framework, under which past real information does not accumulate across
iterations.
Consequently, the recursive estimator cannot fully aggregate information from all
previously observed real samples.
We interpret this phenomenon as an efficiency ceiling induced by limited retention,
rather than an intrinsic limitation of recursive learning itself.

Moreover, when the per-round real sample size $n$ is fixed and the amount of synthetic
data used in each round is sufficiently large, allowing the weight to depend on the
training horizon $T$, i.e., $w = w_T$, enables further improvement of the convergence rate.
Under appropriate choices of $\{w_T\}$, the estimation error admits the rate $O \Big(\frac{\mathrm{tr}(\bm{\Sigma})}{n}\cdot\frac{\log T}{T}\Big)$ which is near-optimal up to a logarithmic factor compared with the benchmark
$O \Big(\frac{\mathrm{tr}(\bm{\Sigma})}{nT}\Big)$ achievable when all past real data are fully retained.
We refer to Section~A.7 in the supplementary file for additional discussion.
\end{remark}

\subsection{Generalized Linear Model}
In this section, we aim to investigate the phenomenon of model collapse within the fresh-data augmentation framework under more general settings, where the results in Section~\ref{SubSec:Gaus2} arise as a special case. The primary technical challenge lies in the absence of a closed-form minimizer at each training step. To address this, we instead analyze the asymptotic behavior of the training process in the recursive training regime.

To ensure the broad applicability of our results, we examine a general class of generalized linear models (GLMs) as case studies. In these models, there is a pair of random variables denoted as $(\bm{X}, \bm{Y})$, where $\bm{Y} \in \mathbb{R}^{d_Y}$ represents a $d_Y$-dimensional response vector and $\bm{X} \in \mathbb{R}^{d_{X} \times d_{\theta}}$ denotes the covariate matrix. We assume the distribution of $\bm{X}$, denoted as $\mathbb{P}_{\bm{X}}$, is known. Given $\bm{X} = \bm{x}$, the density function of the response vector $\bm{Y}$ is given as
\begin{align}
\label{GLM}
    P_{\bm{Y}}(\bm{y} \mid \bm{x}, \bm{\theta}) &= h(\bm{y}) \exp\left\{ (\bm{x}\bm{\theta})^\top S(\bm{y}) - A(\bm{x}\bm{\theta}) \right\}, 
\end{align}
where $\bm{\theta} \in \mathbb{R}^{d_{\theta}}$ is the parameter vector of interest, $S: \mathbb{R}^{d_Y} \to \mathbb{R}^{d_X}$ is the corresponding (vector-valued) sufficient statistic, and the log-partition function $A(\bm{x}\bm{\theta})$ is given as
\begin{align*}
    A(\bm{x}\bm{\theta})
  = \log \int_{\mathbb{R}^{d_Y}}
        \exp \bigl\{(\bm{x}\bm{\theta})^\top  S(\bm{y})\bigr\}
        h(\bm{y})  d\bm{y}.
\end{align*}

The central focus is on the estimation error of the ground-truth parameters during the recursive training process. It is essential to highlight that equation (\ref{GLM}) encompasses several classical statistical models as special cases. Below, we outline a few of these cases (detailed proofs provided in the supplementary file):
\begin{itemize}
    \item[(1)] {Linear Regression}: If \( d_Y = d_X = d \) for some \( d \geq 1 \), \( S(\bm{y}) = \bm{\Sigma}^{-1} \bm{y} \), $A(\bm{x}\bm{\theta})=\frac{1}{2} (\bm{x} \bm{\theta})^\top \bm{\Sigma}^{-1} \bm{x} \bm{\theta}$, then the model simplifies to \( \bm{Y} = \bm{X} \bm{\beta} + \bm{\epsilon} \), with \( \bm{\epsilon} \sim N(\bm{0}, \bm{\Sigma}) \). Particularly, when \( d = 1 \), this reduces to the linear regression model.
    \item[(2)] {Logistic Regression}: If \( d_Y = d_X = d \) for some \( d \geq 1 \), \( S(\bm{y}) = (y_1, \ldots, y_d) \) with \( \bm{y} \in \{0, 1\}^d \), and $A(\bm{x}\bm{\theta})=\sum_{i=1}^d\log(1+\exp(\bm{x}_i\bm{\theta}))$, then we have $P_{\bm{Y}}(\bm{y} \mid \bm{x}, \bm{\theta}) = \prod_{i=1}^d \frac{[\exp(\bm{x}_i \bm{\theta})]^{y_i}}{1 + \exp(\bm{x}_i \bm{\theta})}$, where \( \bm{x}_i \) represents the \( i \)-th row of \( \bm{x} \). Particularly, when \( d = 1 \), this model reduces to the logistic regression.
    \item[(3)] {Poisson Regression}: If \( d_Y = d_X = d \) for some \( d \geq 1 \), \( S(\bm{y}) = \bm{y} \), $A(\bm{x}\bm{\theta})=\sum_{i=1}^d\exp(\bm{x}_i\bm{\theta})$, then we have $P_{\bm{Y}}(\bm{y} \mid \bm{x}, \bm{\theta}) =\prod_{i=1}^d \frac{(\bm{x}_i\bm{\beta})^{y_i}e^{-\bm{x}_i\bm{\beta}}}{y_i!}$. Particularly, when \( d = 1 \), this model reduces to the Poisson regression.
    \item[(4)] {Exponential Family}: If $d_X = d_{\theta} = d$ for some $d \geq 1$ and $\bm{X} = \bm{I}_d$ almost surely, the model in (\ref{GLM}) reduces to the exponential family in the canonical form.
\end{itemize}

From the preceding examples, it is evident that our framework subsumes a broad class of learning problems as special cases. Leveraging this generality, we now shift our focus to examining the phenomenon of model collapse within this unified framework. The overall procedure is outlined in {Fresh Data Augmentation Case}~\ref{alg:recursive_mle}.

\begin{algorithm}[ht]
\caption{Recursive Weighted MLE}
\label{alg:recursive_mle}
\begin{algorithmic}

\STATE {Input:} Initial dataset $\mathcal{D}_0 = \{(\bm{x}_{0,i}, \bm{y}_{0,i})\}_{i=1}^n$, weight parameter $w \in [0,1]$, number of iterations $T$.
\vspace{0.5em}

\STATE {Initialization:} Estimate initial parameter $\widehat{\bm{\theta}}_0$ by {solving the MLE}:
\begin{equation*}
\widehat{\bm{\theta}}_0 
= \arg\min_{\bm{\theta}} \left\{ 
  -\frac{1}{n} \sum_{i=1}^{n} \log P_{\bm{Y}}(\bm{y}_{0,i} \mid \bm{x}_{0,i}, \bm{\theta}) 
\right\}.
\end{equation*}

\FOR{$t = 1,2,\ldots,T$}
    \STATE {1. Data Collection:}
\begin{itemize}
  \item \textit{New Real Data:} Generate a real dataset $\{(\bm{x}_{t,i}, \bm{y}_{t,i})\}_{i=1}^n$ such that 
  $\bm{x}_{t,i} \sim \mathbb{P}_{\bm{X}}$ and $\bm{y}_{t,i} \sim \mathbb{P}_{\bm{Y}}(\cdot \mid \bm{x}_{t,i}, \bm{\theta}^\star)$.

  \item \textit{Synthetic Data:} Generate a synthetic dataset $\{(\widetilde{\bm{x}}_{t,i}, \widetilde{\bm{y}}_{t,i})\}_{i=1}^m$ such that 
  $\widetilde{\bm{x}}_{t,i} \sim \mathbb{P}_{\bm{X}}$ and $\widetilde{\bm{y}}_{t,i} \sim \mathbb{P}_{\bm{Y}}(\cdot \mid \widetilde{\bm{x}}_{t,i}, \widehat{\bm{\theta}}_{t-1}(w,n,m))$.
\end{itemize}
    \STATE {2. Weighted Training:} Update the parameter $\widehat{\bm{\theta}}_t(w,n,m)$ by solving the weighted MLE:
    \begin{equation*}
    \widehat{\bm{\theta}}_t(w,n,m) = \arg\min_{\bm{\theta}} -\left\{
    \frac{w}{n}\sum_{i=1}^{n} \log P_{\bm{Y}}(\bm{y}_{t,i} \mid \bm{x}_{t,i}, \bm{\theta})
    + \frac{1-w}{m} \sum_{i=1}^{m} \log P_{\bm{Y}}(\widetilde{\bm{y}}_{t,i} \mid \widetilde{\bm{x}}_{t,i}, \bm{\theta})
    \right\}.
    \end{equation*}

\ENDFOR

\STATE {Return:} Final estimator $\widehat{\bm{\theta}}_T(w,n,m)$.

\end{algorithmic}
\end{algorithm}

In the initial stage, we observe a real dataset $\mathcal{D}_0 = \{(\bm{x}_{0,i}, \bm{y}_{0,i})\}_{i=1}^n$ and construct an initial estimator $\widehat{\bm{\theta}}_0(w,n,m)$ based on a weighted maximum likelihood procedure:
\begin{align*}
\text{Initial Stage}:
\begin{cases}
   {Initial Real Dataset: } \bm{x}_{0,i} \sim \mathbb{P}_{\bm{X}} \,\,\text{ and }\,\, \bm{y}_{0,i} \sim \mathbb{P}_{\bm{Y}}(\bm{y} \mid \bm{x}_{0,i}, \bm{\theta})  \,\text{ for }\, i \in [n], \\
  {Initial Estimator: }  \widehat{\bm{\theta}}_0 
= \arg\min_{\bm{\theta}} 
  \left\{ 
  -\frac{1}{n} \sum_{i=1}^{n} \log P_{\bm{Y}}(\bm{y}_{0,i} \mid \bm{x}_{0,i}, \bm{\theta}) 
  \right\}.
\end{cases}
\end{align*}
where $\mathbb{P}_{\bm{X}}$ denotes a known covariate distribution.

In the initial stage, we aim to mimic the common practice wherein the first generative model is trained exclusively on real data. This setup reflects standard procedures in the development of language models, where the initial model is typically trained on human-generated text corpora \citep{achiam2023gpt}. Such training on high-quality, authentic data forms the foundation for subsequent model refinement or synthetic data generation.

Subsequently, at the $t$-th step, a synthetic dataset $\widetilde{D}_{t} = \{(\widetilde{\bm{x}}_{t,i}, \widetilde{\bm{y}}_{t,i})\}_{i=1}^m$ is generated using the previously fitted generative model, and a new real dataset $D_t = \{(\bm{x}_{t,i}, \bm{y}_{t,i})\}_{i=1}^n$ is collected from real distribution. To integrate information from both sources, we consider the following weighted negative log-likelihood objective:
\begin{align}
\mathcal{L}_t(\bm{\theta}, w,n,m) = -
\left\{
\frac{w}{n} \sum_{i=1}^{n} \log P_{\bm{Y}}(\bm{y}_{t,i} \mid \bm{x}_{t,i}, \bm{\theta}) +
\frac{1-w}{m} \sum_{i=1}^{m} \log P_{\bm{Y}}(\widetilde{\bm{y}}_{t,i} \mid \widetilde{\bm{x}}_{t,i}, \bm{\theta})
\right\}.
\end{align}
Specifically, the overall procedure at the $t$-th training stage can be summarized as follows.
\begin{align*}
\text{$t$-th Stage}:
\begin{cases}
   \text{New Real Dataset: } \bm{x}_{t,i} \sim \mathbb{P}_{\bm{X}} \,\,\text{ and }\,\, \bm{y}_{t,i} \sim \mathbb{P}_{\bm{Y}}(\bm{y} \mid \bm{x}_{t,i}, \bm{\theta}^\star)  \,\text{ for }\, i \in [n], \\
   \text{Synthetic Dataset: } \widetilde{\bm{x}}_{t,i} \sim \mathbb{P}_{\bm{X}} \,\,\text{ and }\,\, \widetilde{\bm{y}}_{t,i} \sim \mathbb{P}_{\bm{Y}}(\bm{y} \mid \widetilde{\bm{x}}_{t,i}, \widehat{\bm{\theta}}_{t-1}(w,n,m) )  \,\text{ for }\, i \in [m],\\
  \text{$t$-th Estimator: }  \widehat{\bm{\theta}}_t(w,n,m)=\arg \min_{\bm{\theta}} \mathcal{L}_t(\bm{\theta}, w,n,m).
\end{cases}
\end{align*}

At the $t$-th training stage, in addition to the real dataset $D_t$, a synthetic dataset $\widetilde{D}_t$ is also available. This setting is motivated by the practical scenario in which modern generative models are increasingly trained on a mixture of real and synthetic data due to the limited availability of high-quality human-generated content \citep{ding2024data,alemohammad2024selfconsuming}.

\begin{assumption}
\label{Ass:Moment}
Assume that $\mathrm{vec}(\bm{X}) \in \mathbb{R}^{d_{\bm{X}} \times d_{\bm{\theta}}}$ is a sub-Gaussian random vector. That is, there exists a constant $K > 0$ such that, for any $\bm{u} \in \mathbb{R}^{d_{\bm{X}} \times d_{\bm{\theta}}}$ with $\|\bm{u}\|_2 = 1$,
$$
\mathbb{E} \left[\exp \Big(t\,\langle \bm{u},\, \mathrm{vec}(\bm{X}) - \bm \mu \rangle\Big)\right]
\;\le\; \exp \left(\frac{K^2 t^2}{2}\right),
\quad \forall\, t \in \mathbb{R},
$$
where $\bm \mu = \mathbb{E} \big[\mathrm{vec}(\bm{X})\big]$ denotes the mean vector.
\end{assumption}

Assumption \ref{Ass:Moment} imposes a sub-Gaussian tail condition on the vectorized form of $\bm{X}$. This assumption guarantees that the distribution of $\mathrm{vec}(\bm{X})$ exhibits light tails and is generally not restrictive in practice. In particular, when $\bm{X}$ is fixed, Assumption \ref{Ass:Moment} holds trivially.

\begin{assumption}
\label{Ass:ALL}
We assume that the following conditions hold for the model in (\ref{GLM}):
\begin{itemize}
    \item[(1)] $A(\bm{\eta})$ and $\nabla_{\bm{\eta}}^2 A(\bm{\eta})$ are continuous in $\bm{\eta}$ and $S(\bm{Y})$ is measurable in $\bm{Y}$.
    \item[(2)] $\mathbb{E}\,\| S(\bm{Y}) \|_2^2$, $\mathbb{E}\left[\sup_{\bm{\theta} \in \bm{\Theta}} \|  A(\bm{X}\bm{\theta}) \|_2\right]$, and $\mathbb{E}\left[\sup_{\bm{\theta} \in \bm{\Theta}} \| \nabla^2 A(\bm{X}\bm{\theta}) \|_{\mathrm{op}}^2\right]$ are all finite.
    \item[(3)] $\mathbb{E}_{\bm{X} \sim \mathbb{P}_{\bm{X}}} \big[ \bm{X}^\top \nabla^2 A(\bm{X}\bm{\theta}) \bm{X} \big]$ is non-singular at $\bm{\theta} = \bm{\theta}^{\star}$ and $\mathbb{E}[\log P_{\bm{Y}}(\bm{Y}|\bm{X},\bm{\theta})]$ is uniquely maximized by $\bm{\theta}^\star$.
    \item[(4)] There exists $r>0$ such that for any $\widetilde{\bm{\theta}} \in B(\bm{\theta}^\star, r)$ and any $\bm{X}\in\mathbb{R}^{d_X\times d_\theta}$,
\[
  \left|
    \left.
      \frac{\partial^3 A(\bm{\eta})}{\partial \eta_i \partial \eta_j \partial \eta_k}
    \right|_{\bm{\eta} = \mathbf{X}\widetilde{\bm{\theta}}}
  \right|
  \,\le\, h(\mathbf{X}),
  \quad \forall\, 1 \le i,j,k \le d_T,
\]
where $h$ is a non-negative functions wtih finite expectation.
\end{itemize}
\end{assumption}

Assumption \ref{Ass:ALL} summarizes a set of mild regularity conditions for the class of models in (\ref{GLM}) required for our theoretical analysis. 
These conditions are relatively weak and are imposed primarily to ensure basic smoothness properties and distributional regularity of the covariates. They are satisfied by a broad range of commonly used models, including the examples introduced above. For each example, we provide a detailed verification of Assumptions~\ref{Ass:Moment}--\ref{Ass:ALL} in the supplementary file. We emphasize that the purpose of formulating these assumptions is not to restrict the applicability of our framework, but rather to facilitate a transparent asymptotic analysis. In particular, these conditions are sufficiently general to encompass many classical settings while also laying the foundation for extending our results to more complex learning problems.

To assess the estimation accuracy of $\widehat{\bm{\theta}}_t(w,n,m)$ for $\bm{\theta}^\star$, we define the following scaled estimation error for $t \geq 1$:
\begin{align*}
\overline{\mathrm{Err}} \left(\widehat{\bm{\theta}}_t(w,k)\right) 
    \triangleq          \lim_{\substack{n,m\rightarrow\infty \\ \frac{n}{m}\rightarrow k}} n \cdot \mathbb{E} \left[
    \left\Vert \widehat{\bm{\theta}}_t(w,n,m)-\bm{\theta}^\star \right\Vert_2^2
\right].
\end{align*}
As mentioned earlier, deriving a closed-form expression for $\widehat{\bm{\theta}}_t(w,n,m)$ is challenging. Therefore, we focus on analyzing its asymptotic behavior while keeping $k$ fixed. The scaling constant $n$ represents the number of real samples used in each training step.

\begin{theorem}
    \label{Theorem:GLM}
For any $T_0 \in \mathbb{Z}$, under Assumptions \ref{Ass:Moment} and \ref{Ass:ALL}, the following holds for all $t \in [T_0]$
    \begin{align}
\overline{\mathrm{Err}}\left(\widehat{\bm{\theta}}_t(w,k)\right)
    =(1-w)^2 
\overline{\mathrm{Err}}\left(\widehat{\bm{\theta}}_{t-1}(w,k)\right)
    +\Bigl[(1-w)^2k + w^2\Bigr]\textnormal{tr}(\bm\Sigma_0^{-1}),
    \end{align}
    where $\overline{\mathrm{Err}}\left(\widehat{\bm{\theta}}_{0}(w,k)\right)=\textnormal{tr}(\bm{\Sigma}_0^{-1})$ and $\bm{\Sigma}_0 = \mathbb{E}_{\bm{X}\sim \mathbb{P}_{\bm{X}}}\left[\bm{X}^\top \nabla^2 A(\bm{X}\bm{\theta}^\star) \bm{X}\right]$. For this sequence, we can derive that
    \begin{align*}
\overline{\mathrm{Err}}\left(\widehat{\bm{\theta}}_{\infty}(w,k)\right) \triangleq
        \lim_{t\rightarrow \infty}\overline{\mathrm{Err}} \left(\widehat{\bm{\theta}}_t(w,k)\right) = C(w,k)\textnormal{tr}(\bm\Sigma_0^{-1}),
    \end{align*}
    where $C(w,k) =\frac{w^2+(1-w)^2k}{2w-w^2}$. For any $k>0$, $\overline{\mathrm{Err}} \left(\widehat{\bm{\theta}}_{\infty}(w,k)\right)$ attains its minimum at $w^\star = \frac{\sqrt{k^2 + 4k} - k}{2}$, which reduces to the reciprocal of the golden ratio when $k=1$ (i.e., $n=m$).
\end{theorem}

In Theorem~\ref{Theorem:GLM}, we establish a \emph{recursive formula} for the scaled estimation error 
$\overline{\mathrm{Err}} \left(\widehat{\bm{\theta}}_t(w,k)\right)$ for all models under (\ref{GLM}), which provides a systematic characterization of how the estimation error evolves across training steps. Leveraging this recursive relationship, we are able to analyze the asymptotic behavior of the estimation error at each iteration and, in particular, derive its limiting value as $t \to \infty$. Interestingly, within the class of constant-in-time weighting schemes, the asymptotic estimation error 
$\overline{\mathrm{Err}} \left(\widehat{\bm{\theta}}_{\infty}(w,k)\right)$ is also minimized at $ w^\star = \frac{\sqrt{k^2 + 4k} - k}{2}$ for any fixed $k>0$. This result is fully consistent with the findings obtained in the Gaussian case (Section~\ref{SubSec:Gaus2}), thereby demonstrating the robustness of the optimal weighting scheme across different model settings.

\begin{corollary}\label{cor:n_scaled_finite_style}
Under the assumptions of Theorem~\ref{Theorem:GLM}, for any \( t \geq 1 \), we have  
\begin{align*}
    \textnormal{{Optimal Weighting:}} \quad 
    & \overline{\mathrm{Err}} \left(\widehat{\bm{\theta}}_t(w^\star,k)\right)
        = \operatorname{tr} \big(\bm\Sigma_0^{-1}\big)\,
          \Bigl[w^\star + \bigl(1 - w^\star\bigr)^{2t+1}\Bigr], \\
    \textnormal{{Naive Mixing:}} \quad 
    & \overline{\mathrm{Err}} \left(\widehat{\bm{\theta}}_t(w_0,k)\right)
        = \operatorname{tr} \big(\bm\Sigma_0^{-1}\big)\,
          \left[\frac{k+1}{k+2} 
          + \frac{1}{k+2} \left(\frac{1}{k+1}\right)^{2t}\right],
\end{align*}
where \( w^\star = \dfrac{\sqrt{k^2 + 4k} - k}{2} \) denotes the optimal weight, and 
\( w_0 = \dfrac{k}{k+1} \) corresponds to the naive strategy of directly mixing real and synthetic data during recursive training. For any $k >0$ and $t \geq 1$, it holds that $$\overline{\mathrm{Err}} \left(\widehat{\bm{\theta}}_t(w_0,k)\right)>\overline{\mathrm{Err}} \left(\widehat{\bm{\theta}}_t(w^\star,k)\right).$$
This demonstrates the clear advantage of weighted recursive training compared to the naive mixing.

\end{corollary}

From Corollary~\ref{cor:n_scaled_finite_style}, we see that when $T \to \infty$, under optimal weighting the $n$-scaled risk vanishes as $k\to 0$, while under naive merging it remains bounded away from zero. Moreover, the relative loss  $\mathcal{R}_{\infty}(w^\star,n,m)/\mathcal{R}_{\infty}(w_0,n,m)$ tends to zero as $k \to 0$, highlighting that in a practically common scenario where synthetic data vastly outnumber fresh data, adopting the correct weighting scheme is crucial for maintaining efficiency.

\begin{remark}
We note that the comparison in Corollary~\ref{cor:n_scaled_finite_style} focuses on the setting where optimal weighting and naive mixing are evaluated under the same real and synthetic sample sizes. We also discuss the comparison between optimal weighting and naive mixing when the two methods are allowed to use different synthetic sample sizes, with the detailed analysis deferred to Section~A.5 of the supplementary file.
\end{remark}

\begin{corollary}
\label{cor:covariance-op}
Consider the special case of model (\ref{GLM}) where $\bm{X}=\bm{I}$ almost surely and the real distribution $\mathbf{Y}\sim\mathcal{N}(\bm{\mu},\bm{\Sigma})$. Let $\widehat{\bm{\Sigma}}_t(w,n,m)$ be the weighted estimator at the $t$-th iteration under the fresh data augmentation framework \ref{alg:recursive_mle} with ratio $k=n/m$. Then there exists a constant $C_0>0$ (depending only on $\bm{\Sigma}$) such that, as $n\to\infty$, $m\to\infty$ with $n/m\to k$,
\[
\mathbb{E} \left\|\,\sqrt{n}\big(\widehat{\bm{\Sigma}}_T(w,n,m)-\bm{\Sigma}\big)\right\|_{\mathrm{op}}^2
\;=\; \beta_T\,C_0 \;+\; o(1),
\]
where
\[
\beta_T \;=\; (1-w)^{2T} \;+\; \frac{\big((1-w)^2 k + w^2\big)\,\big(1-(1-w)^{2T}\big)}{w(2-w)}.
\]
In particular, letting $T\to\infty$ yields $\lim_{T\to\infty}\,\beta_T \;=\; \frac{(1-w)^2 k + w^2}{\,2w - w^2\,}
\;=\; C(w,k)$.
\end{corollary}

From Corollary~\ref{cor:covariance-op}, we deduce that the result for multivariate Gaussian estimation also applies when the operator norm is used as the metric for covariance matrix estimation. In particular, when $k=1$, the optimal weight that minimizes the expected operator norm of $\sqrt{n}\big(\widehat{\bm{\Sigma}}_T(w,n,m)-\bm{\Sigma}\big)$ is $\tfrac{\sqrt{5}-1}{2}$.

\begin{remark}
Theorem~\ref{Theorem:GLM} assumes the covariate distribution of $\bm X$ is known. Under this assumption, synthetic data at each iteration can be generated by sampling covariates from the same distribution and then producing responses according to the learned conditional model. This assumption is primarily adopted for theoretical clarity. A more realistic scenario is that the covariate distribution is unknown and must also be learned from data. In such cases, one may consider learning a separate generative model for $\bm X$ and jointly updating the marginal model for $\bm X$ and the conditional model for $Y|\bm X$ across iterations. By Theorem~\ref{Theorem:GLM}, when $\bm X$ itself belongs to an exponential family, learning $\bm X$ alone already exhibits the same recursive behavior characterized in our current analysis. This observation suggests that a joint learning framework for $(\bm X, Y)$ may admit similar recursive error characterizations, which we leave as an interesting direction for future work.

\end{remark}

\section{Nonparametric Distribution Estimation}
\label{Sec:EDE}
In this section, we adopt a non-parametric discrete distribution approximation as a case study to investigate the phenomenon of model collapse under the proposed framework. Specifically, we focus on estimating a univariate discrete distribution within a recursive training paradigm, where the generative model is represented by its empirical distribution.  This setting reflects a key characteristic of large language models (LLMs): since an LLM effectively models a high-dimensional discrete distribution over tokens (\cite{bengio2003neural}), repeatedly training on self-generated data can be interpreted as recursively updating an empirical approximation of the underlying true distribution (see Section~A.4 of the supplementary file for further discussion).  By analyzing this simplified non-parametric scenario, we aim to generalize the recursive training results established in Section~\ref{Sec:GLM} to a broader non-parametric context.

At the initial stage, a real dataset $\mathcal{D}_0 = \{x_{0,i}\}_{i=1}^n$ is sampled from the underlying real distribution, which is characterized by its cumulative distribution function (CDF) $F(\cdot)$ and probability density function (PDF) $p(x)$. Using $\mathcal{D}_0$, a probability mass function (PMF) can be constructed as  
\[
p_0(x) = \frac{1}{n} \sum_{i=1}^n \bm{1}_{\{x = x_{0,i}\}},
\]  
where $\bm{1}_{\{x = x_{0,i}\}}$ is the indicator function, equal to 1 if $x = x_{0,i}$ and 0 otherwise. This empirical PMF serves for generating synthetic data for the subsequent training step. 

At the $t$-th step, a new real dataset $\mathcal{D}_t = \{x_{t,i}\}_{i=1}^n$ emerges and a synthetic dataset $\widetilde{\mathcal{D}}_t = \{\widetilde{x}_{t,i}\}_{i=1}^m$ is generated based on $p_{t-1}(x)$ via sampling. To construct the generative model at the $t$-th step, we propose a weighted combination of the PMFs derived from the synthetic dataset $\widetilde{\mathcal{D}}_t$ and the real dataset $\mathcal{D}_t$ as follows:  
\begin{align}
    \label{GeneModel}
p_t(x|w,n,m) = \frac{(1-w)}{m} \sum_{i=1}^m \bm{1}_{\{x = \widetilde{x}_{t,i}\}}
        +  \frac{w}{n} \sum_{i=1}^n \bm{1}_{\{x = x_{t,i}\}}, 
\end{align}  
It is worth noting that when $w = \frac{n}{n+m}$, the generative model $p_t(x)$ simplifies to the PMF constructed directly from the combined dataset $\mathcal{D}_t \cup \widetilde{\mathcal{D}}_t$.  

To evaluate the proximity of $p_t(x)$ to $p(x)$, we consider the corresponding empirical CDF of $p_t(x)$, defined as
\begin{align*}
    F_t(x|w,n,m) =  \frac{(1-w)}{m} \sum_{i=1}^m \bm{1}_{\{ \widetilde{x}_{t,i} \leq x\}}
        +  \frac{w}{n} \sum_{i=1}^n \bm{1}_{\{x_{t,i} \leq x\}}. 
\end{align*}
Then, we adopt the Cram\'er--von Mises $\omega^2$ criterion \citep{anderson1962distribution,csorgHo1996exact} to evaluate the estimation error of $F_t(\cdot \mid w, n, m)$:
\begin{align*}
\mathrm{Err}(F_t(\cdot|w,n,m))=\int_{\mathbb{R}} \left[F_t(x|w,n,m)-F(x)\right]^2 dF(x).
\end{align*}

The general procedure of the above discussion is summarized in {Fresh Data Augmentation Framework}~\ref{alg:empirical_pmf}, which outputs the empirical CDF after $T$ steps of recursive estimation. In this problem, we are still interested in the limiting error of the final CDF, i.e., $\mathrm{Err}(F_T(x \mid w, n, m))$ as $T$ approaches infinity.

\begin{algorithm}[h]
\caption{- Empirical Probability Mass Function Estimation}
\label{alg:empirical_pmf}
\begin{algorithmic}
\STATE {Initialization:} Compute initial distribution estimation $p_0(x)$ using the empirical probability mass function based on dataset $\mathcal{D}_0 = \{x_{0,i}\}_{i=1}^n$:
\begin{equation*}
    F_0(x) = \frac{1}{n} \sum_{i=1}^n \bm{1}_{\{x_{0,i} \leq x \}},
\end{equation*}
where $\bm{1}_{\{x = x_{0,i}\}}$ is the indicator function, which equals 1 if $x = x_{0,i}$ and 0 otherwise.

\FOR{$t = 1,2,\ldots,T$}
    \STATE {1. Data Collection:}
    \begin{itemize}
        \item Sample synthetic dataset $\widetilde{\mathcal{D}}_t = \{\widetilde{x}_{t,i}\}_{i=1}^m$ from the previous probability mass function $p_{t-1}(x|w,n,m)$ of $F_{t-1}(x|w,n,m)$
        \item Collect real dataset $\mathcal{D}_t = \{x_{t,i}\}_{i=1}^n$ from the real distribution.
    \end{itemize}

    \STATE {2. Distribution Update:}
    Compute the updated distribution estimation $p_t(x)$ as a weighted combination of empirical distributions based on synthetic and real datasets:
    \begin{equation*}
        F_t(x|w,n,m) =  \frac{1-w}{m} \sum_{i=1}^m \bm{1}_{\{\widetilde{x}_{t,i}\leq  x  \}}
        +  \frac{w}{n} \sum_{i=1}^n \bm{1}_{\{x_{t,i} \leq x  \}},
    \end{equation*}
    where $w \in [0,1]$ is the weighting parameter.
\ENDFOR

\RETURN $F_T(x|w,n,m)$.
\end{algorithmic}
\end{algorithm}

\begin{theorem}
Assume that $p(x)$ is a continuous density function, then for any $w \in (0, 1]$ and $n, m \geq 2$, it holds that  
\label{thm:nonparametric}
\begin{align}
\label{Res_FInal}
\mathrm{Err}(F_{\infty}(\cdot|w,n,m))\triangleq\lim_{t \to \infty} \mathrm{Err}(F_t(\cdot|w,n,m)) =
\
\frac{1}{6} \cdot \frac{\frac{w^2}{n} + \frac{(1-w)^2}{m}}{1 - (1-w)^2 \left( 1 - \frac{1}{m} \right)},
\end{align}
where $w^\star = \frac{\sqrt{n^{2} + \left(4m - 2\right) n - 4m + 5} - n + 3}{2m + 2}$ is the optimal weight. Particularly, when $m,n\rightarrow \infty$ with $\frac{n}{m}=k$, $w^\star \to \frac{\sqrt{k^2+4k}-k}{2}$. 
\end{theorem}

In Theorem \ref{thm:nonparametric}, we establish the limiting error of $F_{T}(\cdot \mid w, n, m)$ in estimating $F(\cdot)$ under the Cramér--von Mises $\omega^2$ criterion. Several notable insights can be derived from (\ref{Res_FInal}). First, the error $\mathrm{Err}(F_{\infty}(\cdot \mid w, n, m))$ remains finite for all values of $(w, n, m)$, a result that directly follows from the finiteness of the Cramér--von Mises $\omega^2$ criterion. Notably, unlike previous results, as $w$ approaches $0$ (the fully synthetic case), $\mathrm{Err}(F_{\infty}(\cdot \mid w, n, m))$ converges to the fixed value $\frac{1}{6}$, independent of the sample sizes $n$ and $m$. Secondly, the behavior of $\mathrm{Err}(F_{\infty}(\cdot \mid w, n, m))$ can be categorized into two main stages, depending on the choice of $w$:
\begin{itemize}
    \item {Stage 1 (No Improvement Stage)}: When $0<w \leq \frac{n - 1}{n + 2m - 1}$, we have 
    $$
    \frac{\mathrm{Err}\big(F_{\infty}(\cdot \mid w, n, m)\big)}{\mathrm{Err}(F_{\infty}(\cdot \mid 1, n, m))} \geq 1,
    $$
    meaning that incorporating synthetic data downgrades estimation efficiency, offering no improvement.
    \item {Stage 2 (Improvement Stage)}: When $w > \frac{n - 1}{n + 2m - 1}$, we have 
    $$
    \frac{\mathrm{Err}\big(F_{\infty}(\cdot \mid w, n, m)\big)}{\mathrm{Err}(F_{\infty}(\cdot \mid 1, n, m))} < 1,
    $$
    indicating that incorporating synthetic data improves estimation efficiency compared to using only the newly emerging real data.
\end{itemize}

Moreover, note that as $m \to \infty$, the no-improvement stage gradually disappears because the term $\frac{n - 1}{n + 2m - 1}$ converges to zero. This implies that, provided a sufficient amount of synthetic data is generated at each training step, incorporating synthetic data with a non-negligible weight $w$ consistently reduces the estimation error relative to using only real data. Furthermore, the optimal weight $w^\star$ in the non-parametric distribution estimation converges to the same asymptotic value, namely, $w^\star = \frac{\sqrt{k^2 + 4k} - k}{2}$.

\section{Experiments}
\label{Sec:Exp}
In this section, we present a series of simulations and a real-world application to validate our theoretical findings. Our objectives are threefold. First, we demonstrate that model collapse occurs in most recursive training processes, resulting in exploding estimation errors. Second, we show the existence of an optimal weight for weighted training, which corresponds to the reciprocal of the golden ratio when real and synthetic data are equally present. Third, we highlight that, compared to the naive approach of mixing real and synthetic data for training, the weighting schemes lead to improved model performance\footnote{The code for all experiments is available at \url{https://github.com/MukaiCodes/Model_Collapse_Codes.git}.}.

\subsection{Simulations}

\begin{figure}[h!]
    \centering
    % First row
    \begin{subfigure}[b]{0.322\textwidth}
        \centering
        \includegraphics[width=\textwidth]{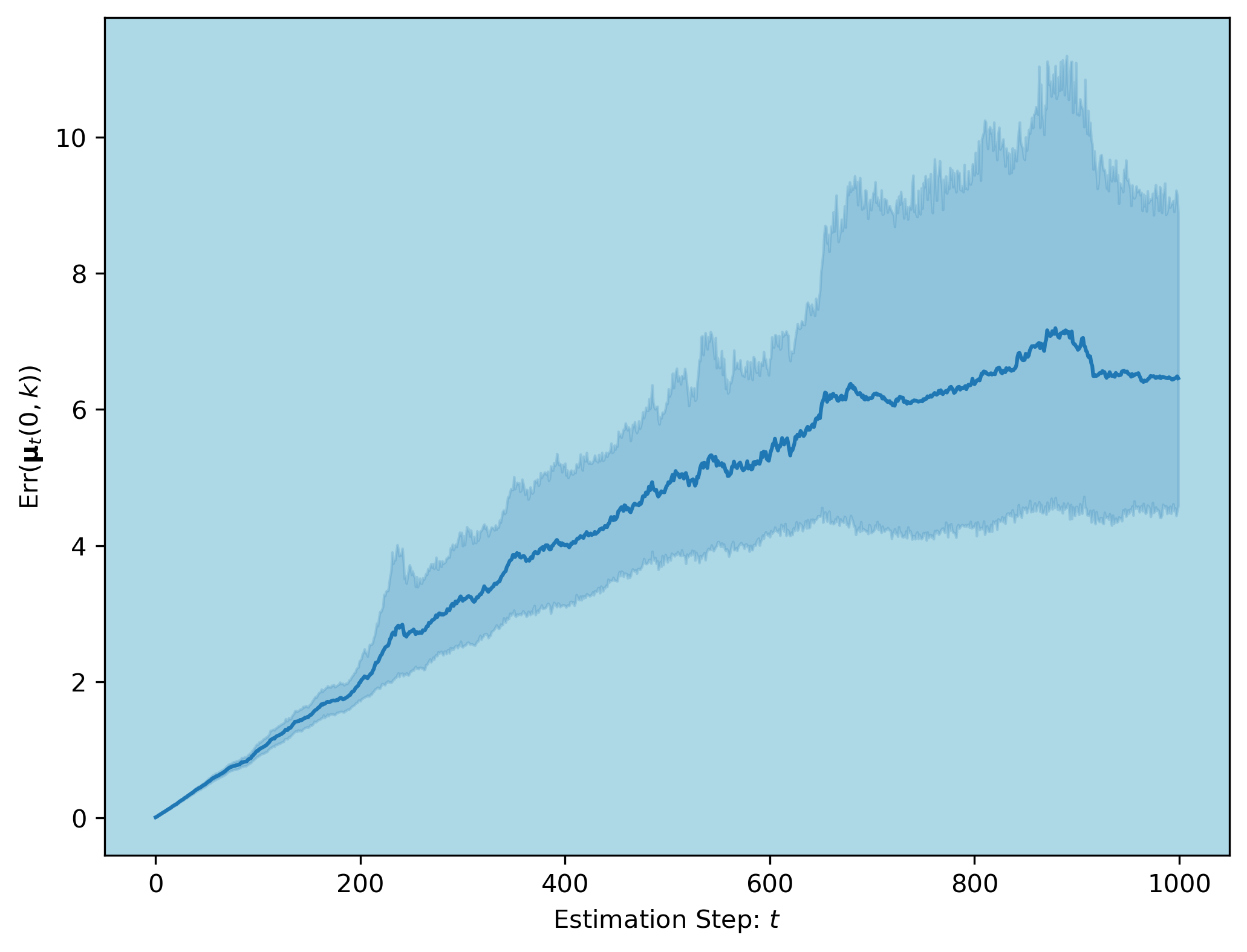}
        \caption{Gaussian Mean}
    \end{subfigure}
    % Second row
    \begin{subfigure}[b]{0.322\textwidth}
        \centering
        \includegraphics[width=\textwidth]{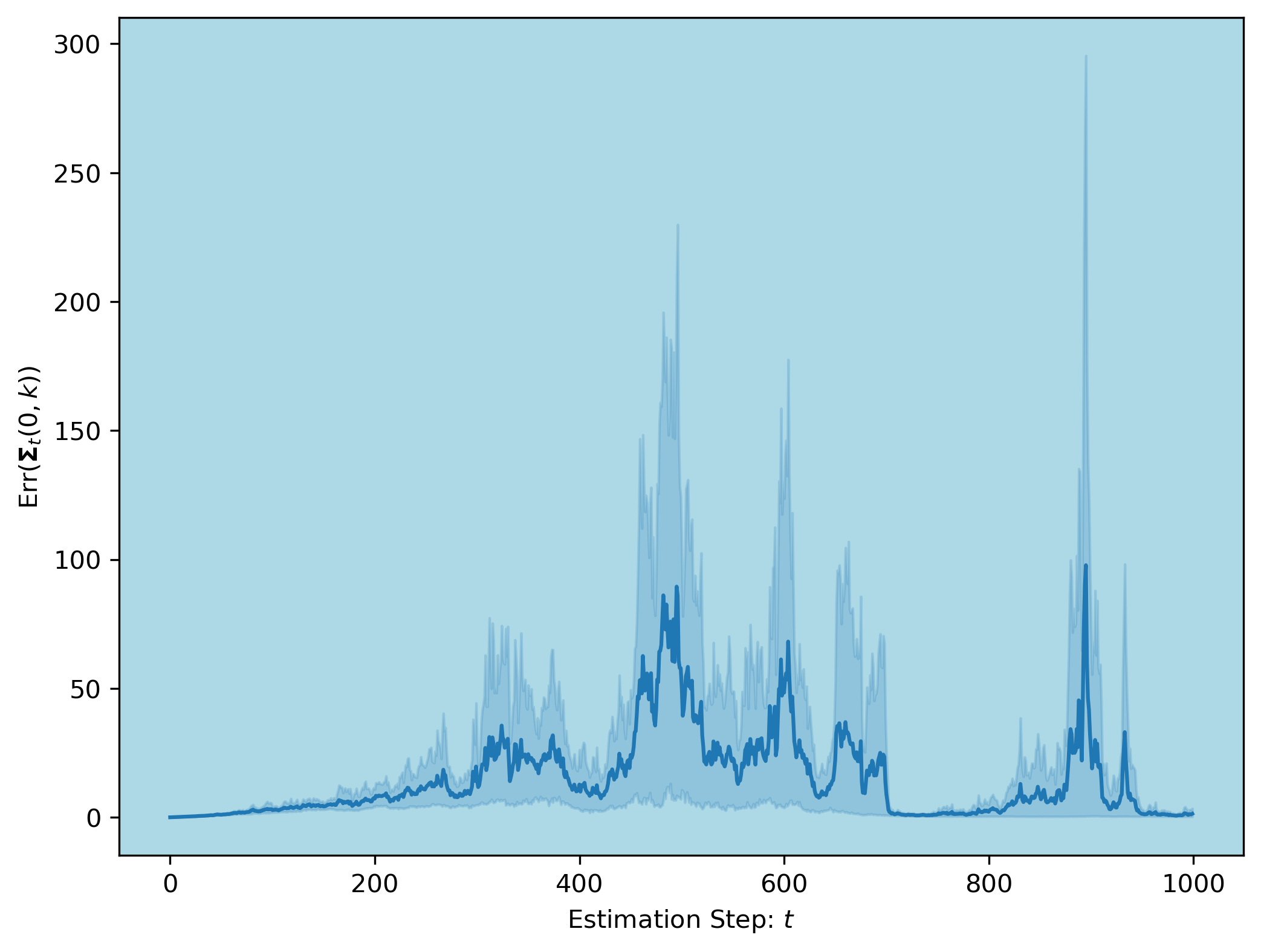}
        \caption{Gaussian Variance}
    \end{subfigure}
    \begin{subfigure}[b]{0.322\textwidth}
        \centering
        \includegraphics[width=\textwidth]{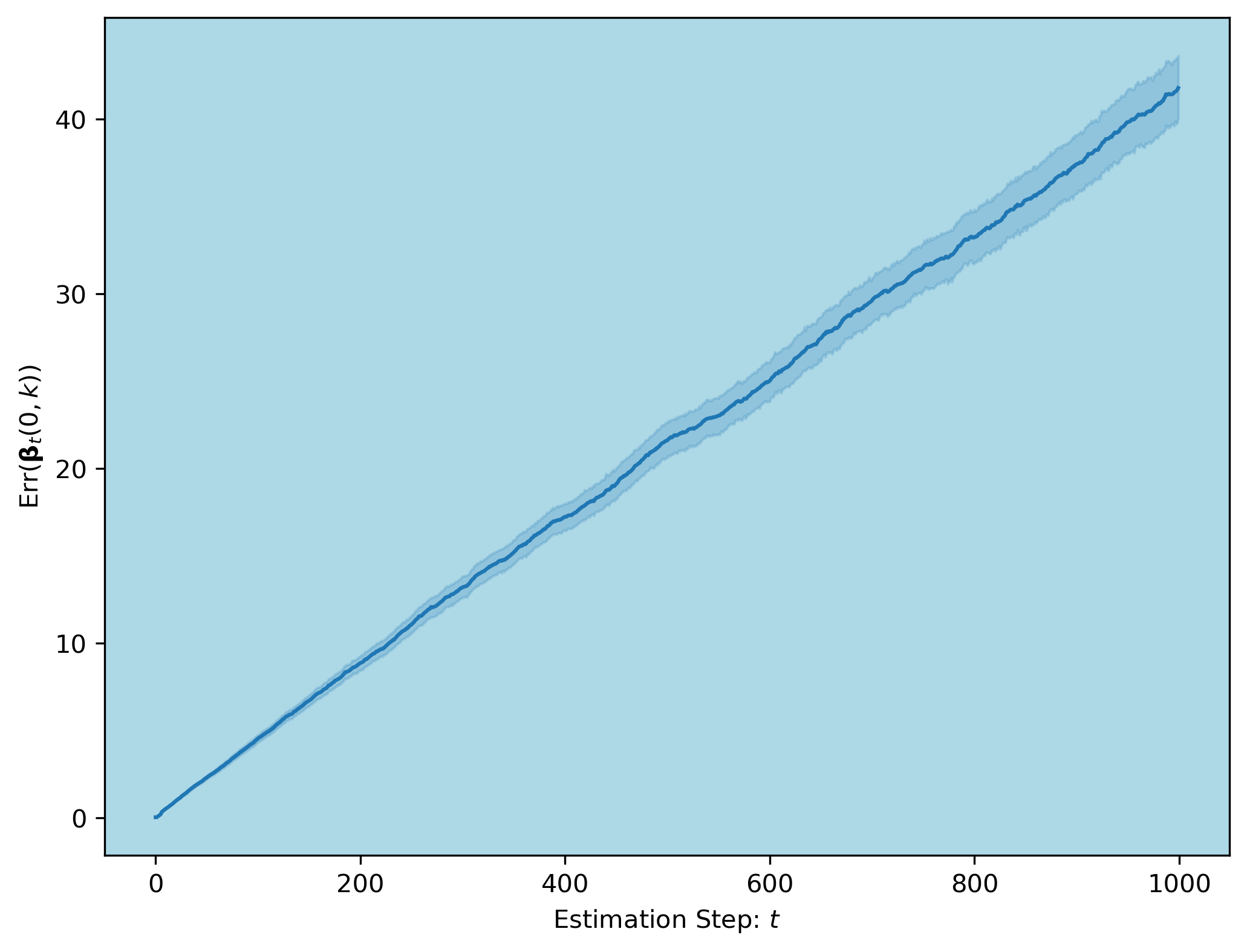}
        \caption{Linear Regression}
    \end{subfigure}
        \begin{subfigure}[b]{0.322\textwidth}
        \centering
        \includegraphics[width=\textwidth]{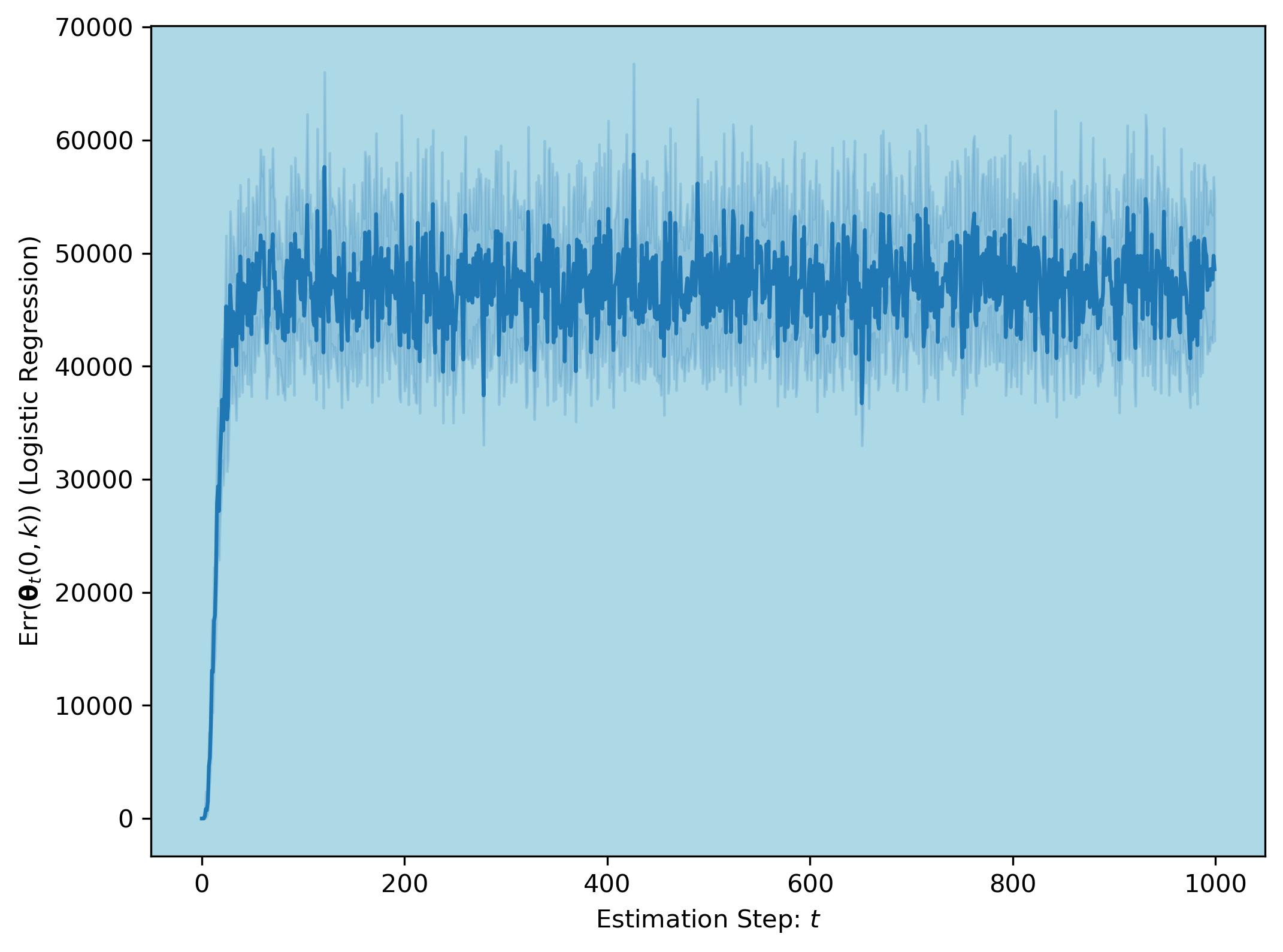}
        \caption{Logistic Regression}
    \end{subfigure}
    % Second row
    \begin{subfigure}[b]{0.322\textwidth}
        \centering
        \includegraphics[width=\textwidth]{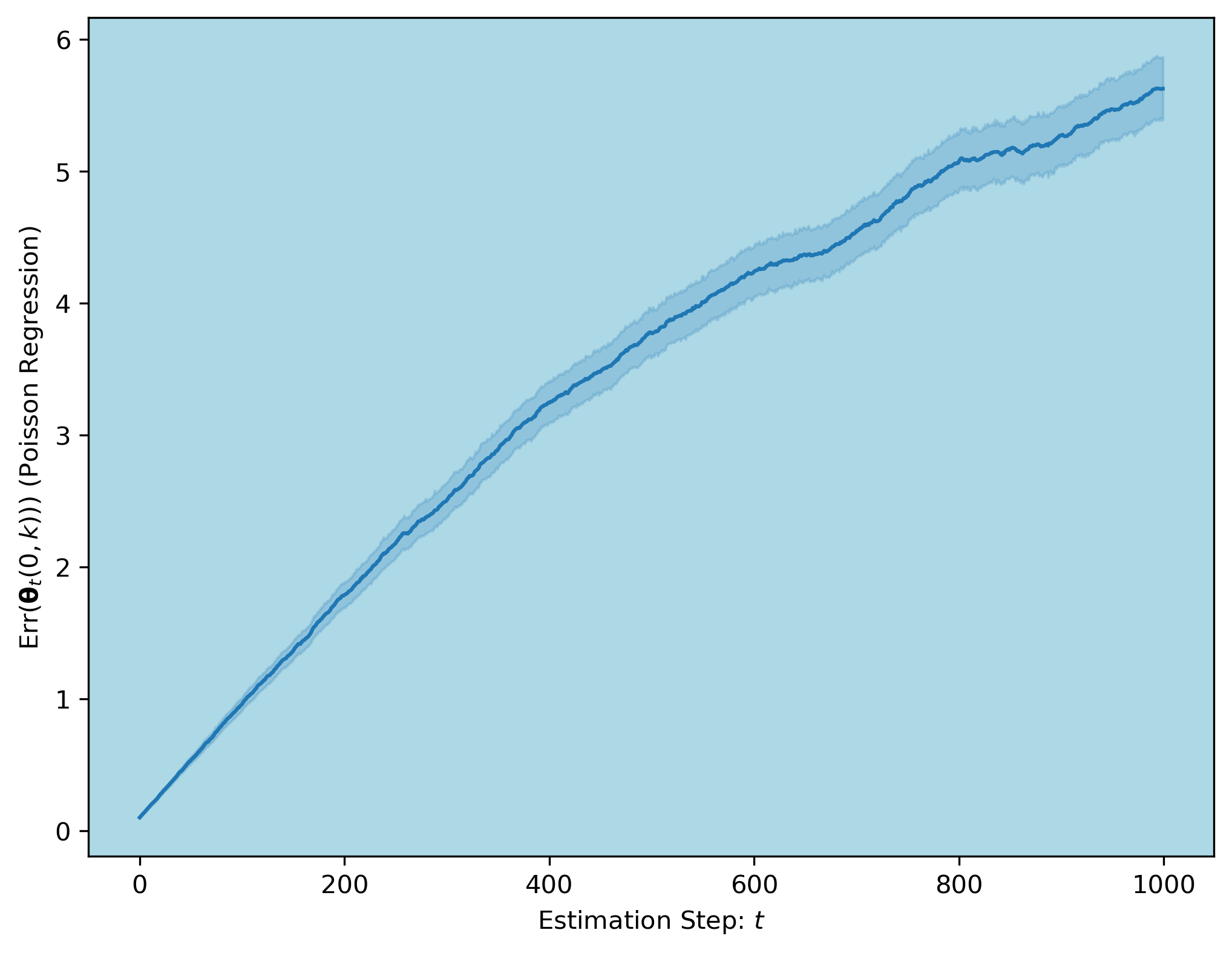}
        \caption{Poisson Regression}
    \end{subfigure}
    \begin{subfigure}[b]{0.322\textwidth}
        \centering
        \includegraphics[width=\textwidth]{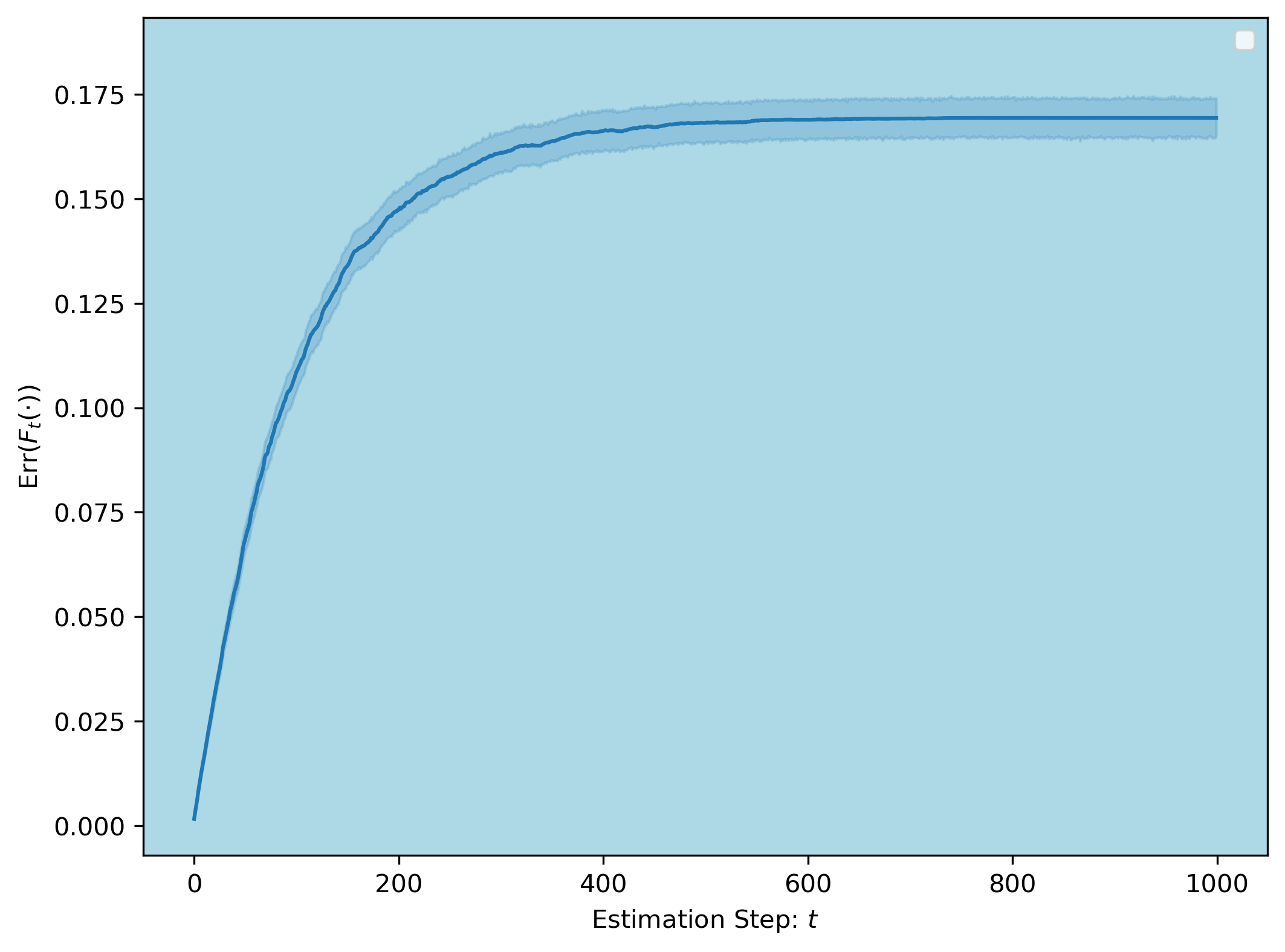}
        \caption{CDF}
    \end{subfigure}
    \caption{The averaged values of estimation errors of all models over $1,000$ replications during the first $1,000$ steps in the fully synthetic case $(w=0)$ are reported for Scenario I (Fully Synthetic Case).
    }
    \label{fig:S1_FS}
\end{figure}

\noindent
{Scenario 1 (Model Collapse in Fully Synthetic Case):} 
In this scenario, we investigate the impact of recursive training on parameter estimation across several commonly used parametric models. Specifically, we consider Gaussian mean and covariance estimation, fixed-design and random-design linear regression, logistic regression, Poisson regression, and cumulative distribution function (CDF) estimation. For the simulations, we generate \( n = 100 \) samples with \( p = 4 \) covariates $\mathbf{X}$ generated from $\mathcal{N}\left( \mathbf{0}, \frac{1}{p} \mathbf{I}_p \right)$. For the regression models, the true regression coefficients are set as $\boldsymbol{\theta} = (1, 1, \dots, 1)^\top \in \mathbb{R}^p$. Each model is recursively trained for \( T = 1,000 \) steps, and the entire experiment is repeated $1,000$ times. At each step, we track the averaged estimation errors. The experimental results are reported in Figure \ref{fig:S1_FS}.

Figure \ref{fig:S1_FS} reports the averaged estimation errors across all models over \(1,000\) replications during the first \(1,000\) recursive training steps under the fully synthetic setting (\(w=0\)). For Gaussian mean estimation, \(\mathrm{Err}(\bm{\mu}_t(0,n,m))\) increases gradually with \(t\), and the 95\% confidence intervals for \(\Vert \bm{\mu}_t(0,n,m) - \bm{\mu} \Vert_2^2\) widen accordingly. As \(t \to \infty\), \(\mathrm{Err}(\bm{\mu}_t(w,n,m))\) diverges, consistent with Corollary~\ref{Coro:Multi_Mean}. Covariance estimation diverges even faster, in line with Corollary~\ref{Coro:Multi_Mean}. Similar patterns hold for logistic regression, linear regression and Poisson regression, and CDF estimation: recursive training without weighting (\(w=0\)) leads to steadily increasing errors and eventual divergence. For non-parametric CDF estimation, the error stabilizes at \(1/6\) due to the boundedness of the Cram\'er--von Mises \(\omega^2\) criterion.

\begin{figure}[h!]
    \centering
    % First row
    \begin{subfigure}[b]{0.322\textwidth}
        \centering
        \includegraphics[width=\textwidth]{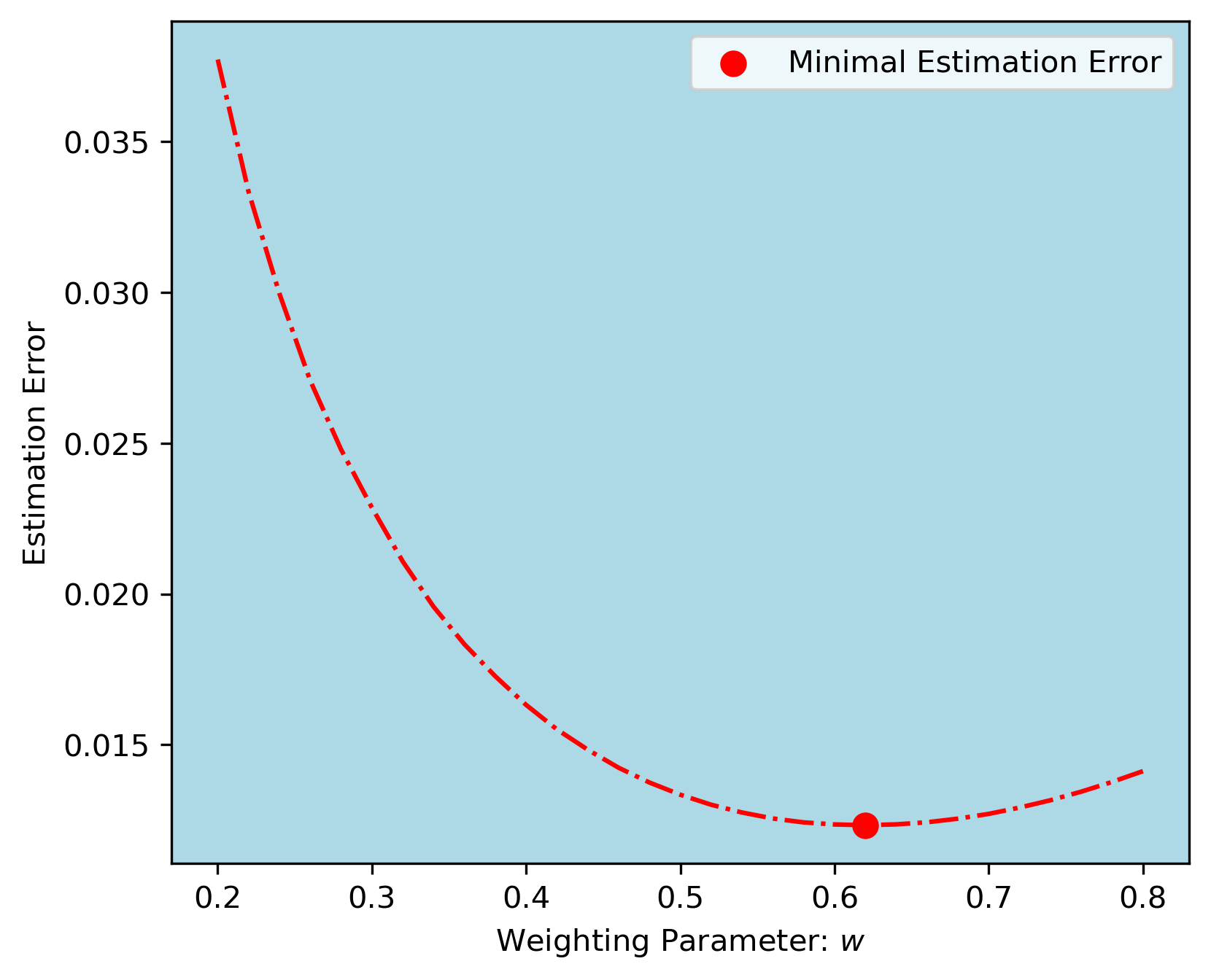}
        \caption{Gaussian Mean}
    \end{subfigure}
    % Second row
    \begin{subfigure}[b]{0.322\textwidth}
        \centering
        \includegraphics[width=\textwidth]{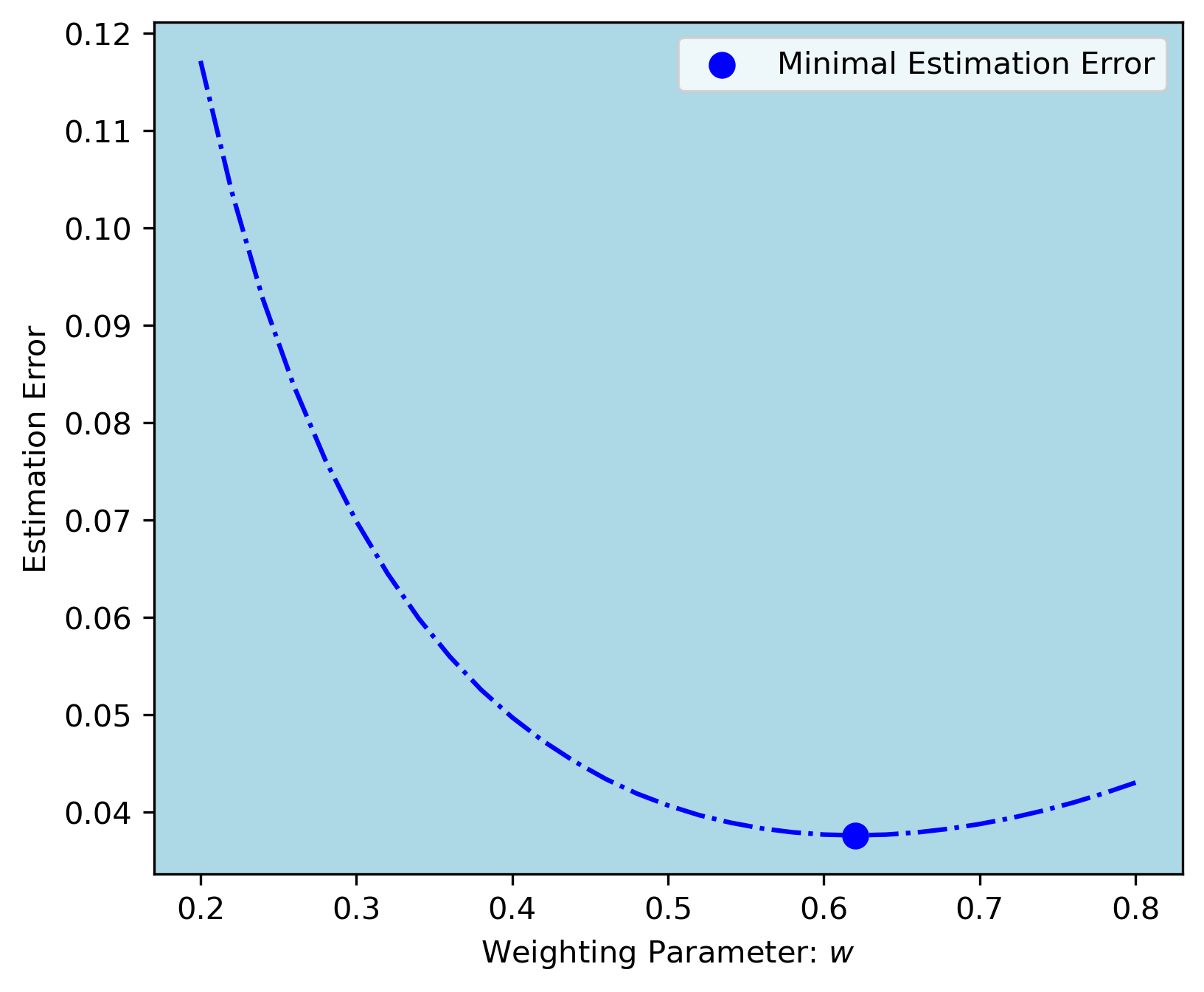}
        \caption{Gaussian Variance}
    \end{subfigure}
    \begin{subfigure}[b]{0.322\textwidth}
        \centering
        \includegraphics[width=\textwidth]{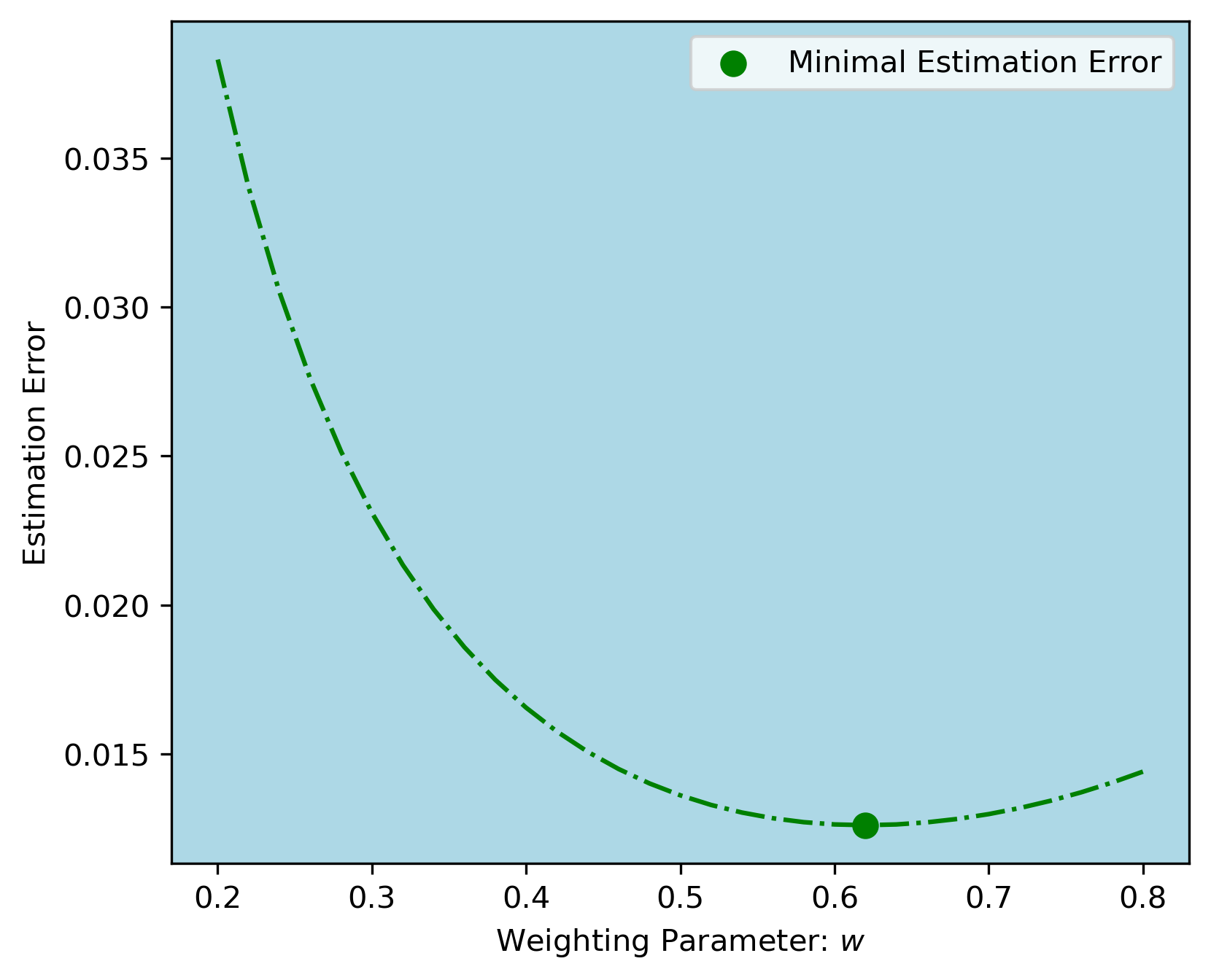}
        \caption{Linear Regression}
    \end{subfigure}
        \begin{subfigure}[b]{0.322\textwidth}
        \centering
        \includegraphics[width=\textwidth]{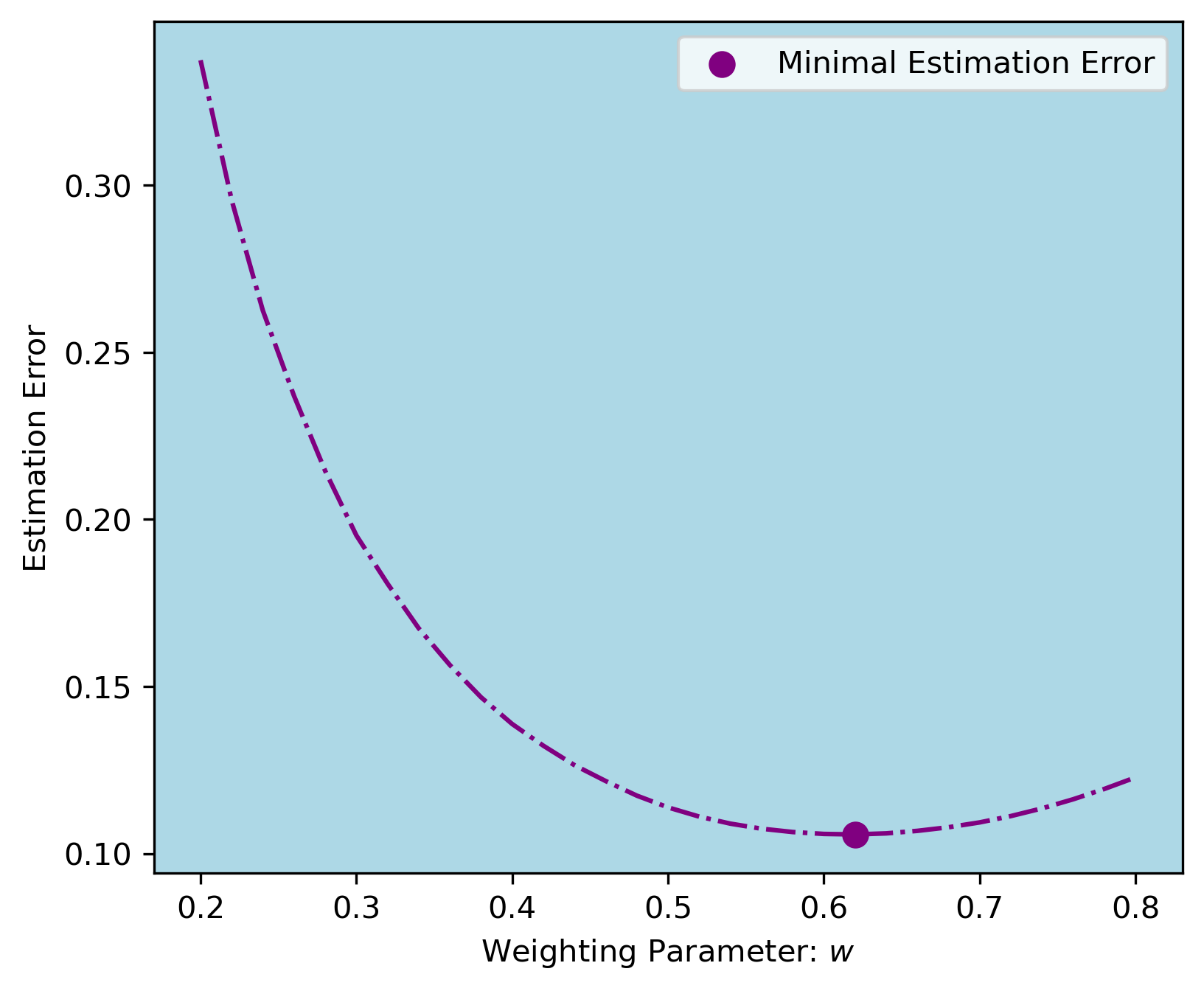}
        \caption{Logistic Regression}
    \end{subfigure}
    % Second row
    \begin{subfigure}[b]{0.322\textwidth}
        \centering
        \includegraphics[width=\textwidth]{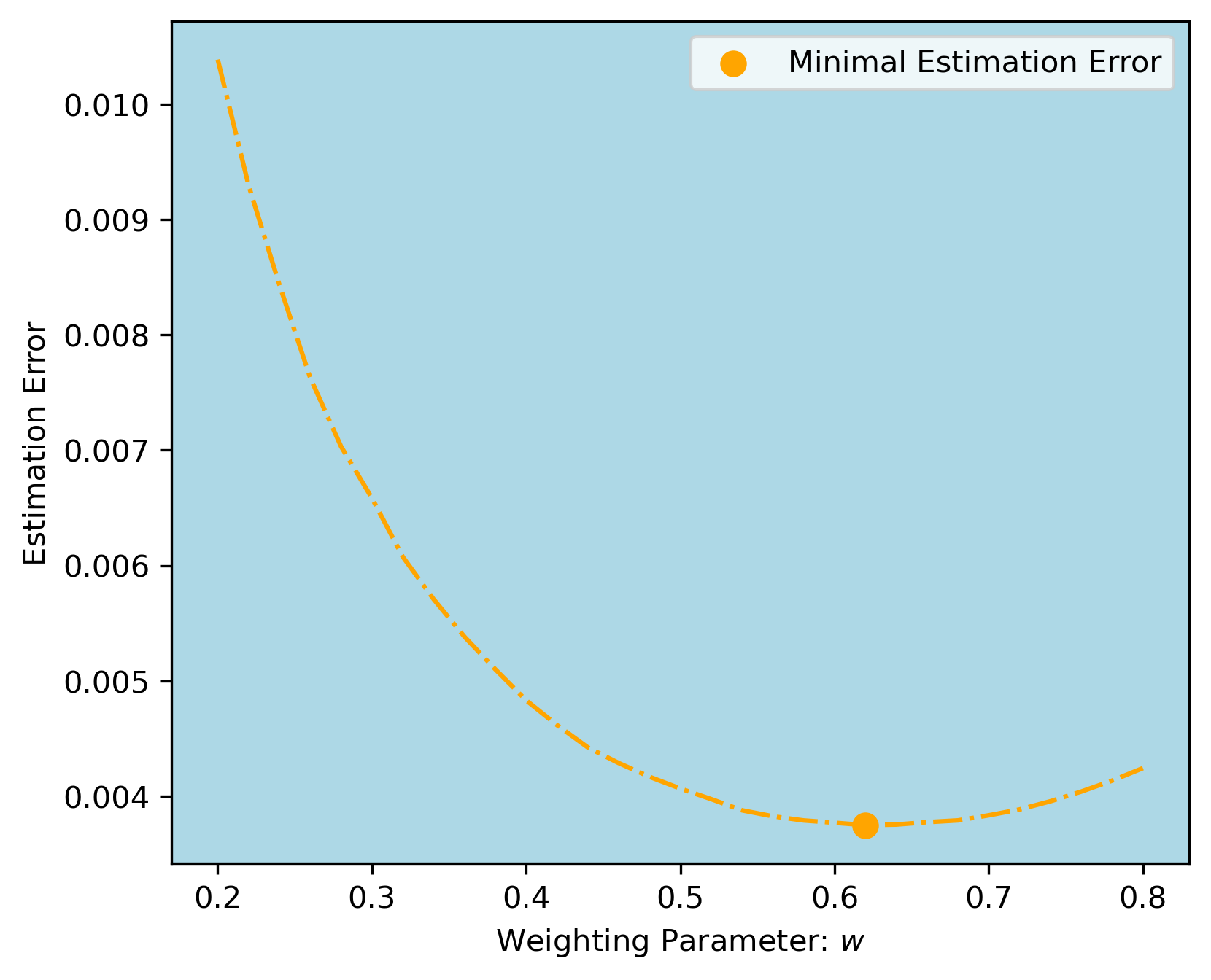}
        \caption{Poisson Regression}
    \end{subfigure}
    \begin{subfigure}[b]{0.322\textwidth}
        \centering
        \includegraphics[width=\textwidth]{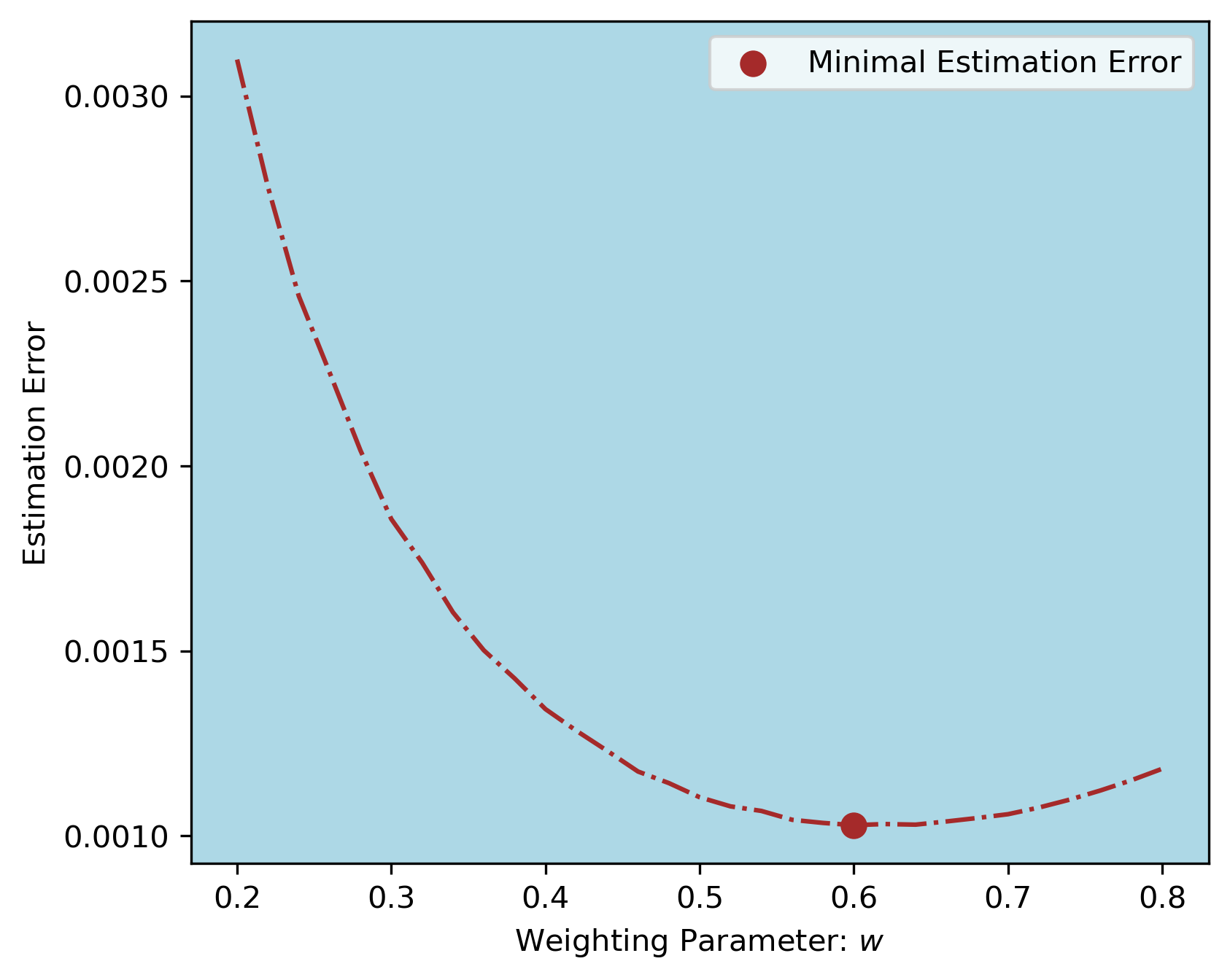}
        \caption{CDF}
    \end{subfigure}
    \caption{The averaged values of estimation errors under different values of $w$ with $(n,T,k) = (100,10^3,1)$ over 1,000 replications. The minimum values of estimation errors are highlighted, which are close to $\frac{\sqrt{5}-1}{2}$.}
    \label{fig:S2_comparison}
\end{figure}

\noindent
{Scenario 2 (Golden Ratio Verification):} In this scenario, we aim to validate our theoretical findings, which suggest that when $n = m$, the optimal proportion of synthetic data corresponds to the reciprocal of the golden ratio. To achieve this, we adopt the same experimental setup for simulated data generation as outlined in Scenario 1, with $w \in \{0.2 + 0.02 \times i \mid i \in [30]\}$ and $(n, T, k) = (100, 200, 1)$. To estimate the limiting estimation errors, we use the average of the last 100 estimation errors as an approximation. For instance, in Gaussian mean estimation, we approximate $\mathrm{Err}(\bm{\mu}_{\infty}(w,n,m))$ using  
\begin{align}
\label{LE_esti}
    \mathrm{Err}(\bm{\mu}_{\infty}(w,n,m)) \approx
\frac{1}{50} \sum_{t=T-50}^{T} \mathrm{Err}(\bm{\mu}_t(w,n,m)).
\end{align} 
Each case is repeated 2,000 times, and the average estimated limiting errors, along with their 95\% confidence intervals, are reported in Figure \ref{fig:S2_comparison}.

The results in Figure~\ref{fig:S2_comparison} are consistent with our theoretical findings. When $n = m$ ($k = 1$), the optimal weighting parameter that minimizes all estimation errors is approximately $0.62$, as shown in Figure~\ref{fig:S2_comparison}. Interestingly, this value is the closest, within the experimental range $w \in \{0.2 + 0.02 \times i \mid i \in [30]\}$, to the reciprocal of the golden ratio, i.e., $\frac{\sqrt{5}-1}{2}$. This suggests that, during recursive training, the optimal weight assigned to real data should be approximately the reciprocal of the golden ratio.

\begin{figure}[h!]
    \centering
    % First row
    \begin{subfigure}[b]{0.322\textwidth}
        \centering
        \includegraphics[width=\textwidth]{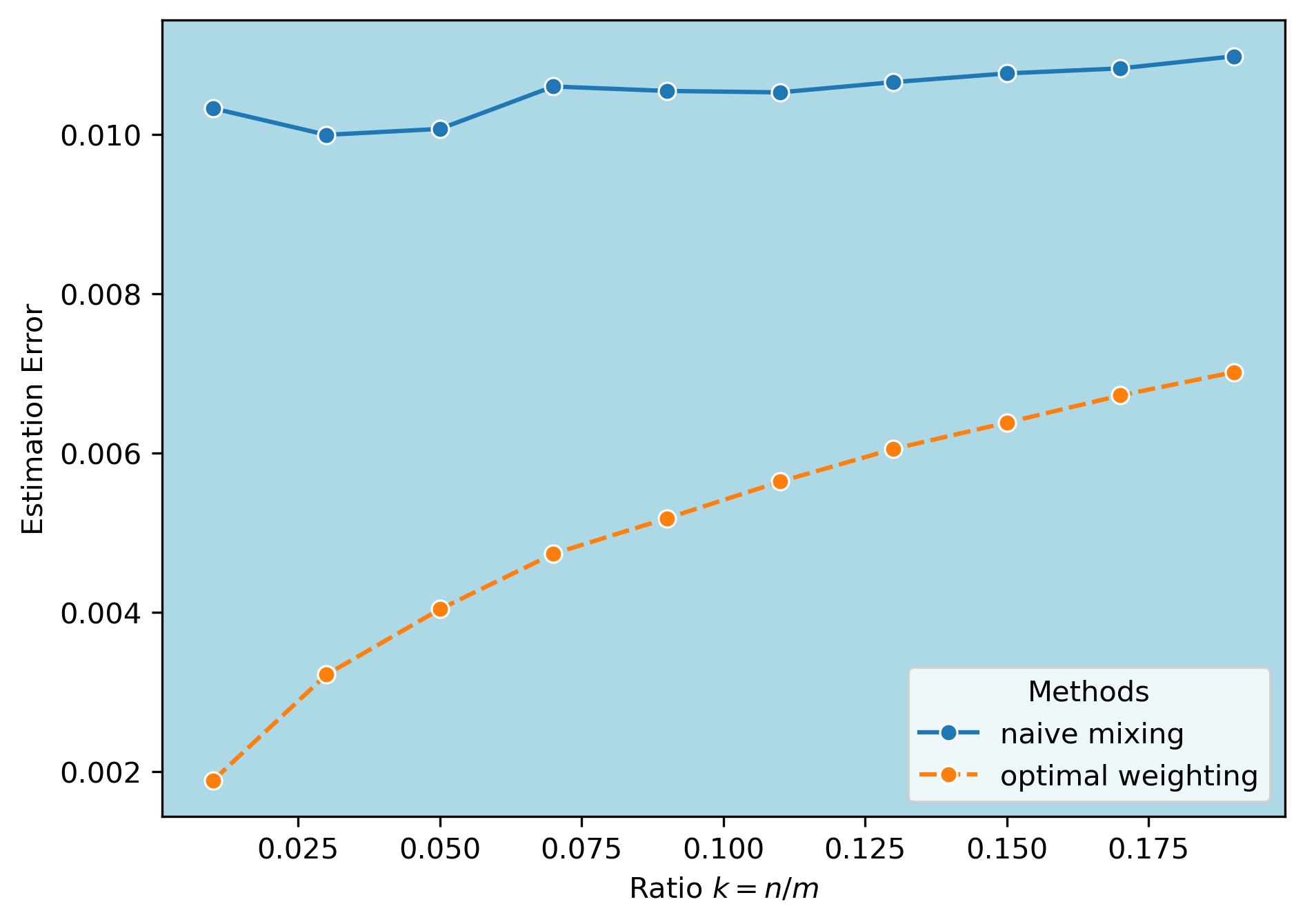}
        \caption{Gaussian Mean}
    \end{subfigure}
    % Second row
    \begin{subfigure}[b]{0.322\textwidth}
        \centering
        \includegraphics[width=\textwidth]{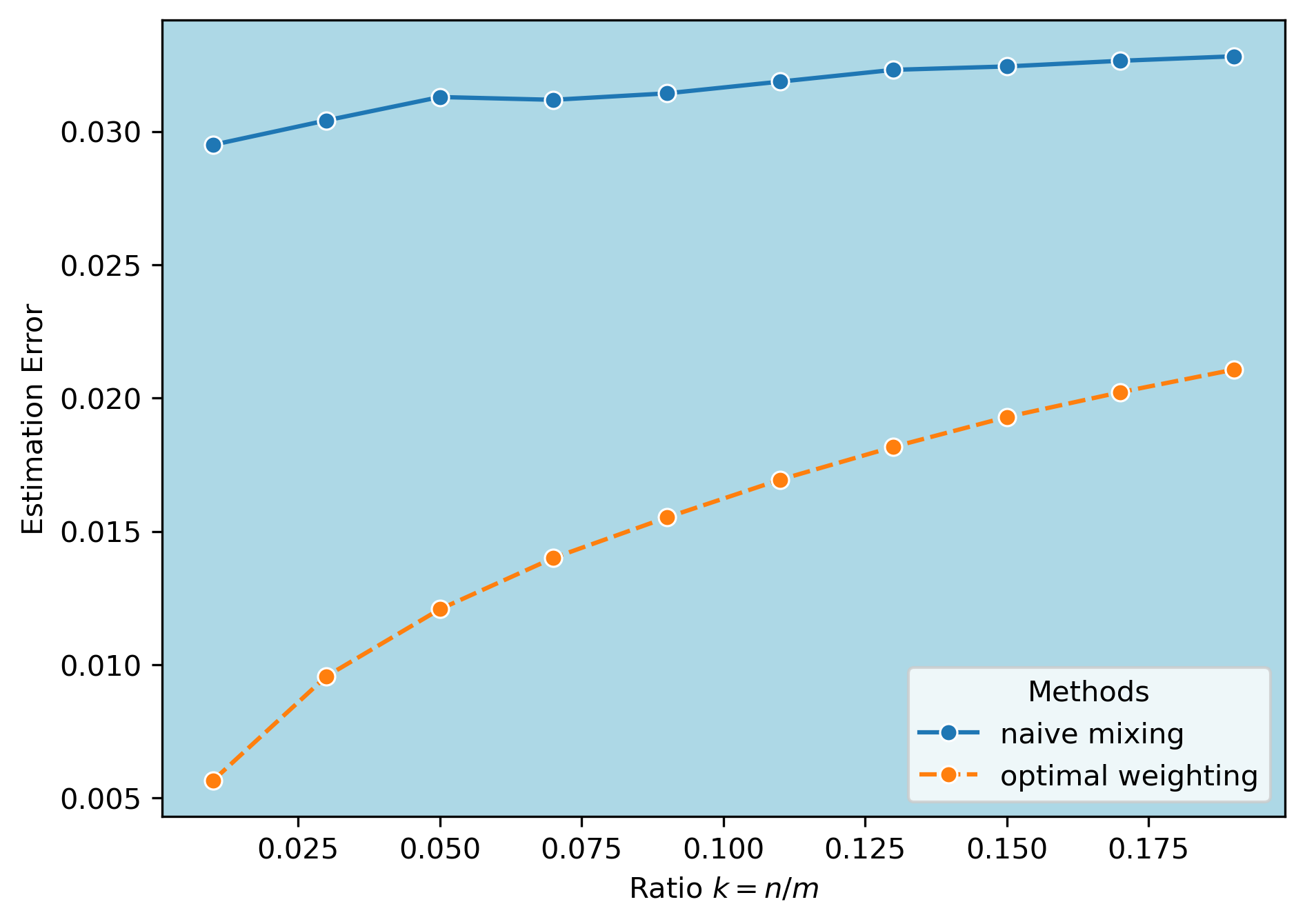}
        \caption{Gaussian Variance}
    \end{subfigure}
    \begin{subfigure}[b]{0.322\textwidth}
        \centering
        \includegraphics[width=\textwidth]{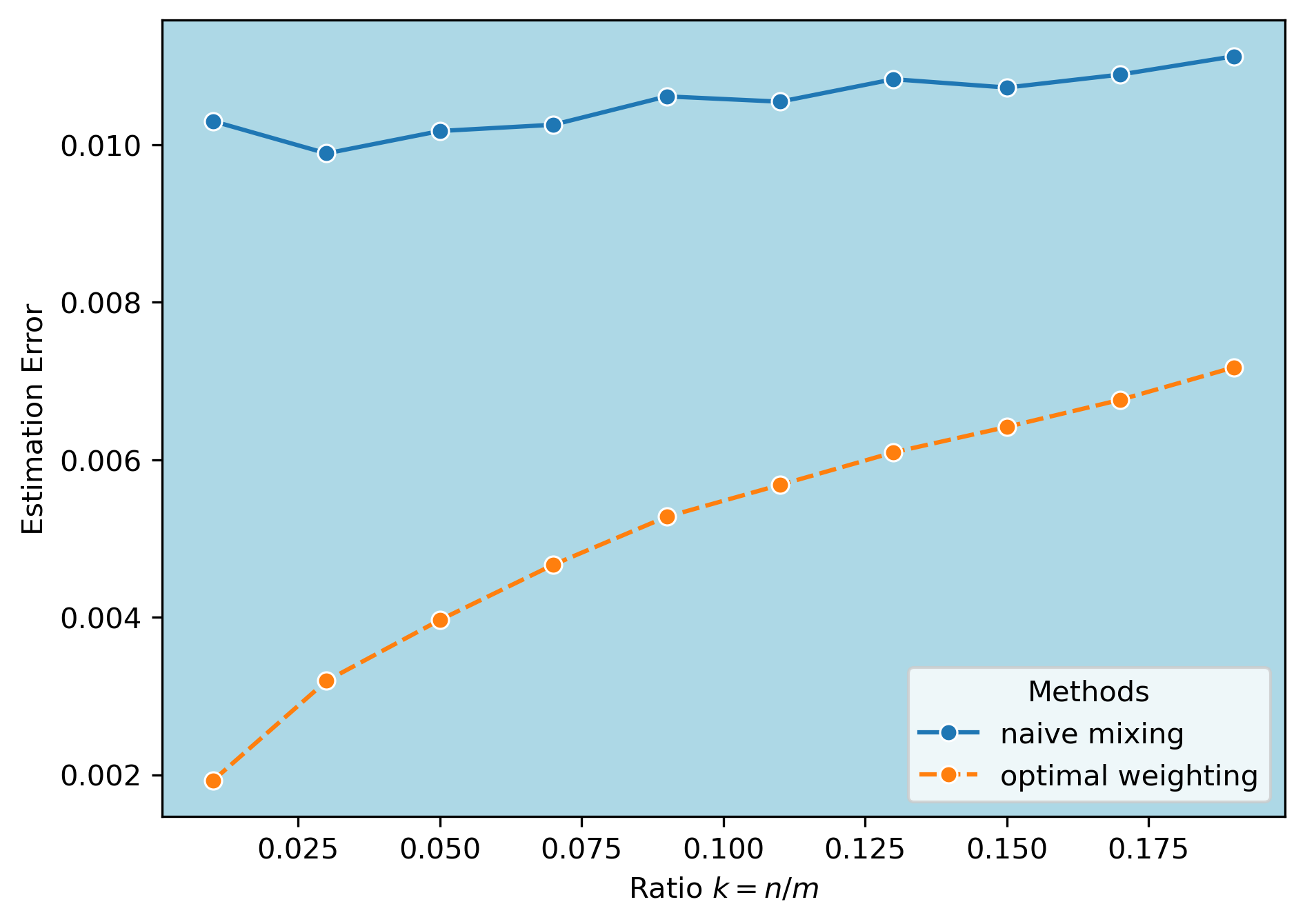}
        \caption{Linear Regression}
    \end{subfigure}
        \begin{subfigure}[b]{0.322\textwidth}
        \centering
        \includegraphics[width=\textwidth]{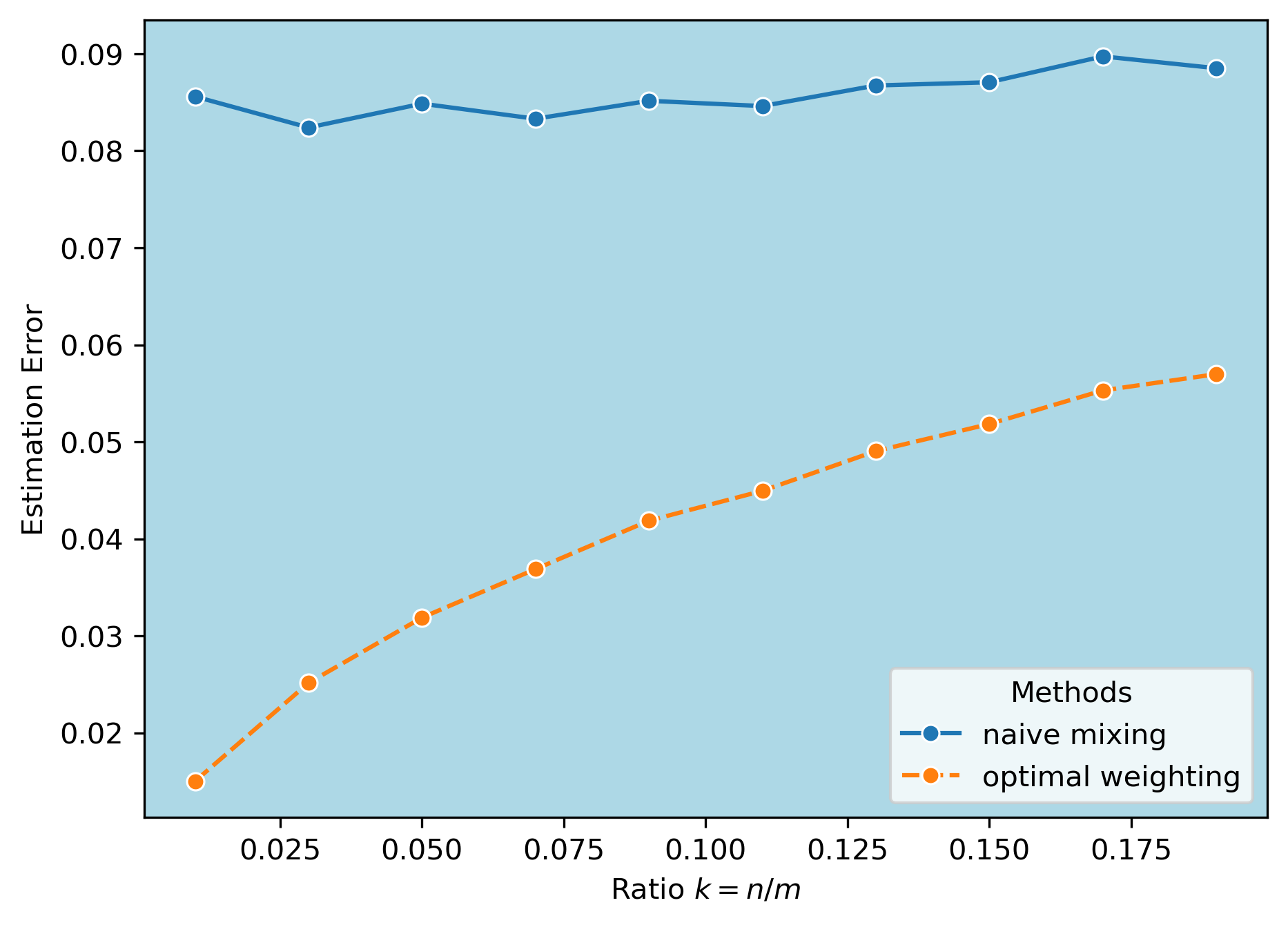}
        \caption{Logistic Regression}
    \end{subfigure}
    % Second row
    \begin{subfigure}[b]{0.322\textwidth}
        \centering
        \includegraphics[width=\textwidth]{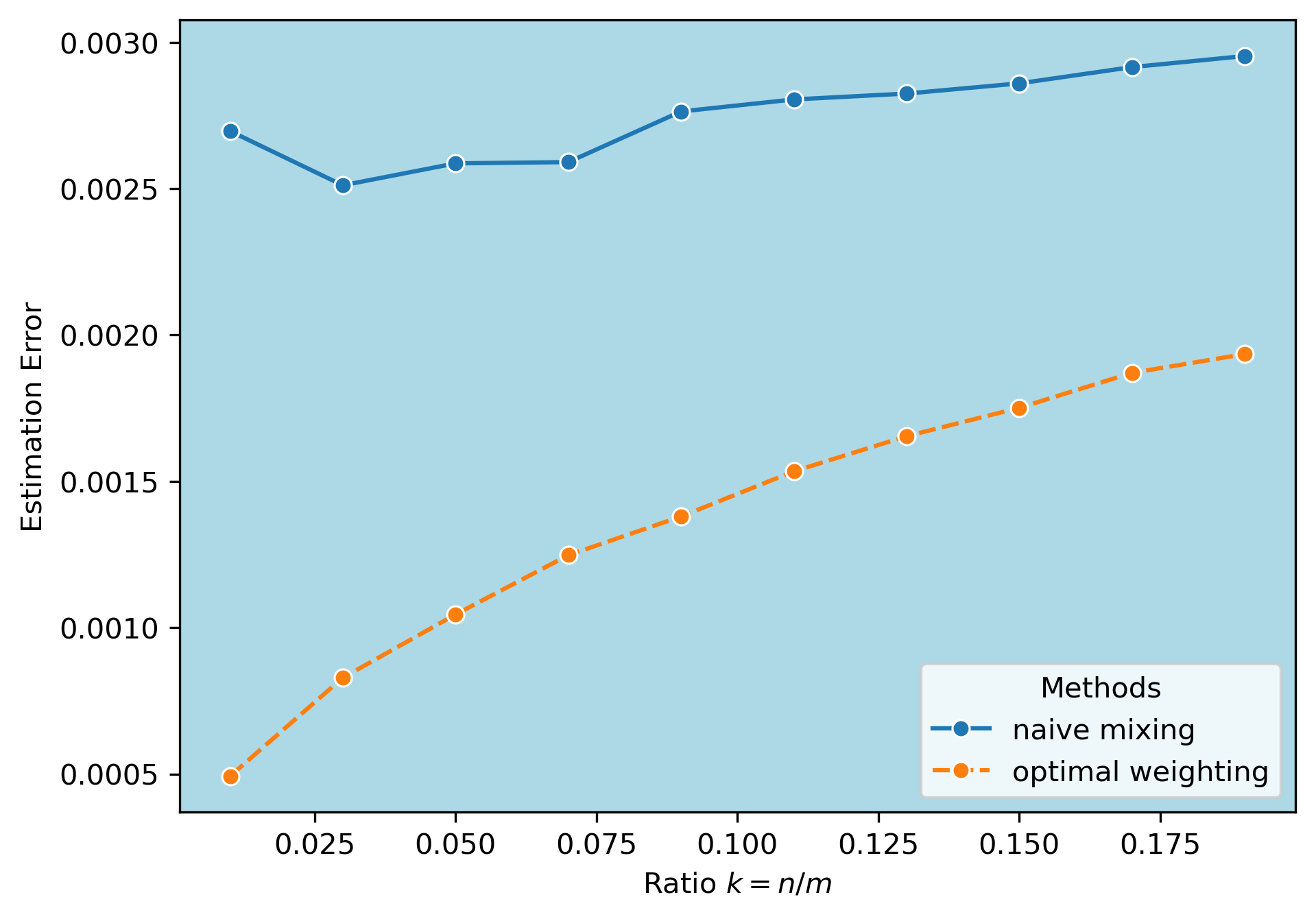}
        \caption{Poisson Regression}
    \end{subfigure}
    \begin{subfigure}[b]{0.322\textwidth}
        \centering
        \includegraphics[width=\textwidth]{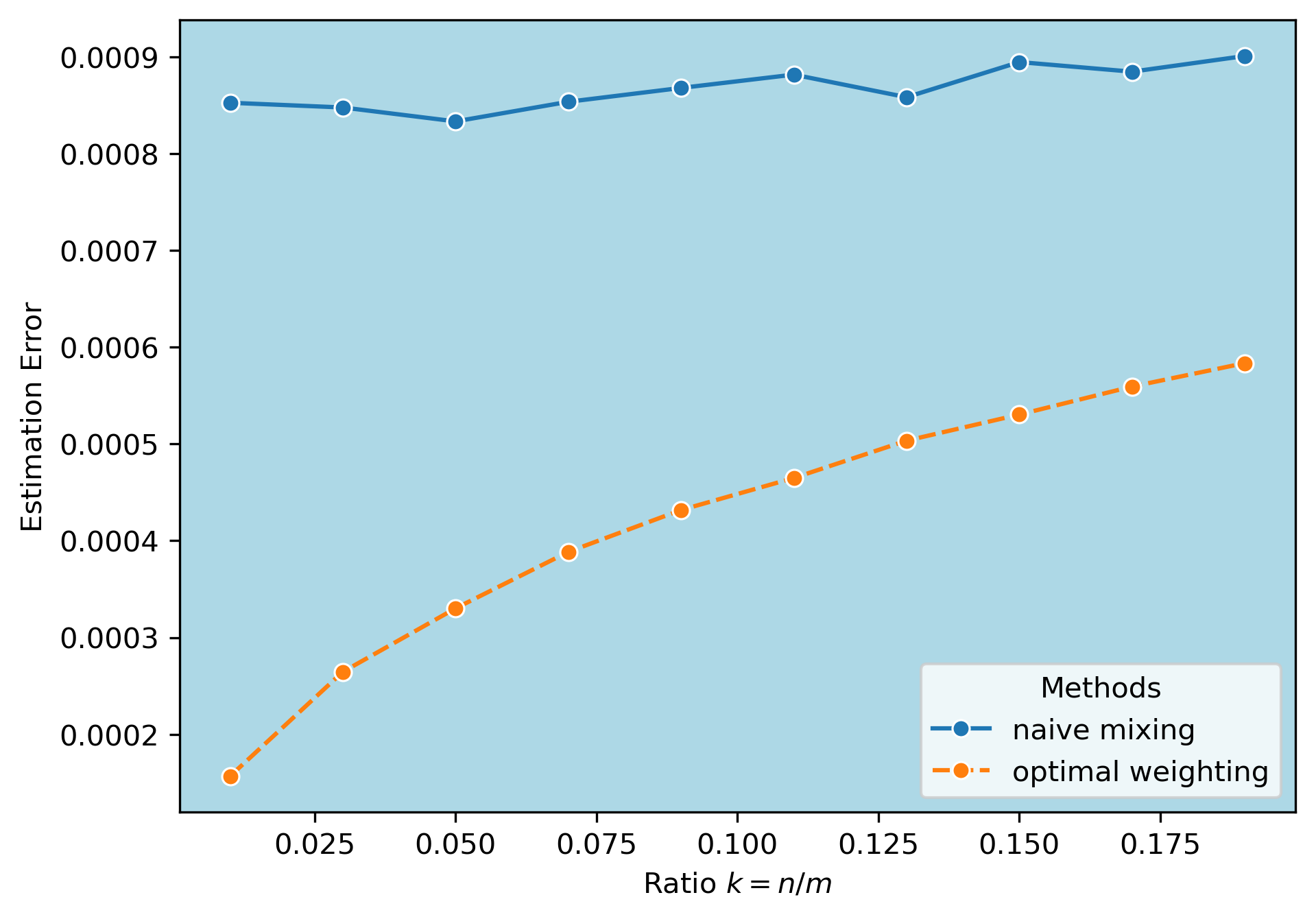}
        \caption{CDF}
    \end{subfigure}
    \caption{The averaged estimation errors of the proposed recursive weighted training scheme and the naive mixed training scheme across different values of $k$, with $(n, T) = (100, 10^3)$, based on 1,000 replications.}
    \label{fig:S3_comparison}
\end{figure}

\noindent
{Scenario 3 (Weighted Training Outperforms Direct Mixing):} In this scenario, we aim to demonstrate that synthetic data can effectively reduce estimation errors by leveraging the optimal weighting parameter $w^\star$ within the fresh data augmentation framework, compared with naively mixing real and synthetic data for training. To illustrate this, we adopt the same models and simulated data generation setup as in Scenarios 1 and 2, fixing $(n, T) = (100, 200)$. We vary the proportion of synthetic data per round, considering $k \in \{0.01 + 0.02 \times i : 0 \leq i \leq 9\}$. For the proposed method, we adopt the optimal weighting parameter $w^\star = \frac{\sqrt{k^2+4k}-k}{2}$ Each experiment is repeated 1,000 times, and the averaged estimation errors are reported in Figure~\ref{fig:S3_comparison}.

As shown in Figure~\ref{fig:S3_comparison}, the proposed method consistently outperforms the naive mixed training approach across all model setups. Consistent with our theoretical results, the estimation error decreases as $k$ increases, indicating that the proposed framework effectively leverages previous real data by integrating a weighted training scheme with synthetic data. In contrast, the naive mixed training method shows limited performance gains when more additional synthetic data are introduced.

\subsection{Real Application}
In this section, we use the Adult dataset \citep{adult_2} to empirically validate our theoretical findings on model collapse. The dataset consists of 48,842 instances, each with 14 features used to predict whether an individual’s annual income exceeds $50K$. These features include continuous variables such as capital-gain, age, and hours-per-week, which, although rounded to integers, maintain their continuous nature. Additionally, the dataset contains categorical variables, including education level, race, and others.

In this application, we consider two main tasks including recursive logistic regression and nonparametric distribution approximation. For the logistic regression, after removing rows with missing values, we extract five key numerical features---\textit{age}, \textit{education\_num}, \textit{capital\_gain}, \textit{capital\_loss}, and \textit{hours\_per\_week}---and transform the income variable into a binary label indicating whether an individual earns more than \$50K annually. For the nonparametric distribution approximation, we consider \textit{workclass}, \textit{education}, and \textit{marital\_status}.

\begin{figure}[h!]
    \centering
    % First row
    % Second row
    \begin{subfigure}[b]{0.464\textwidth}
        \centering
        \includegraphics[width=\textwidth]{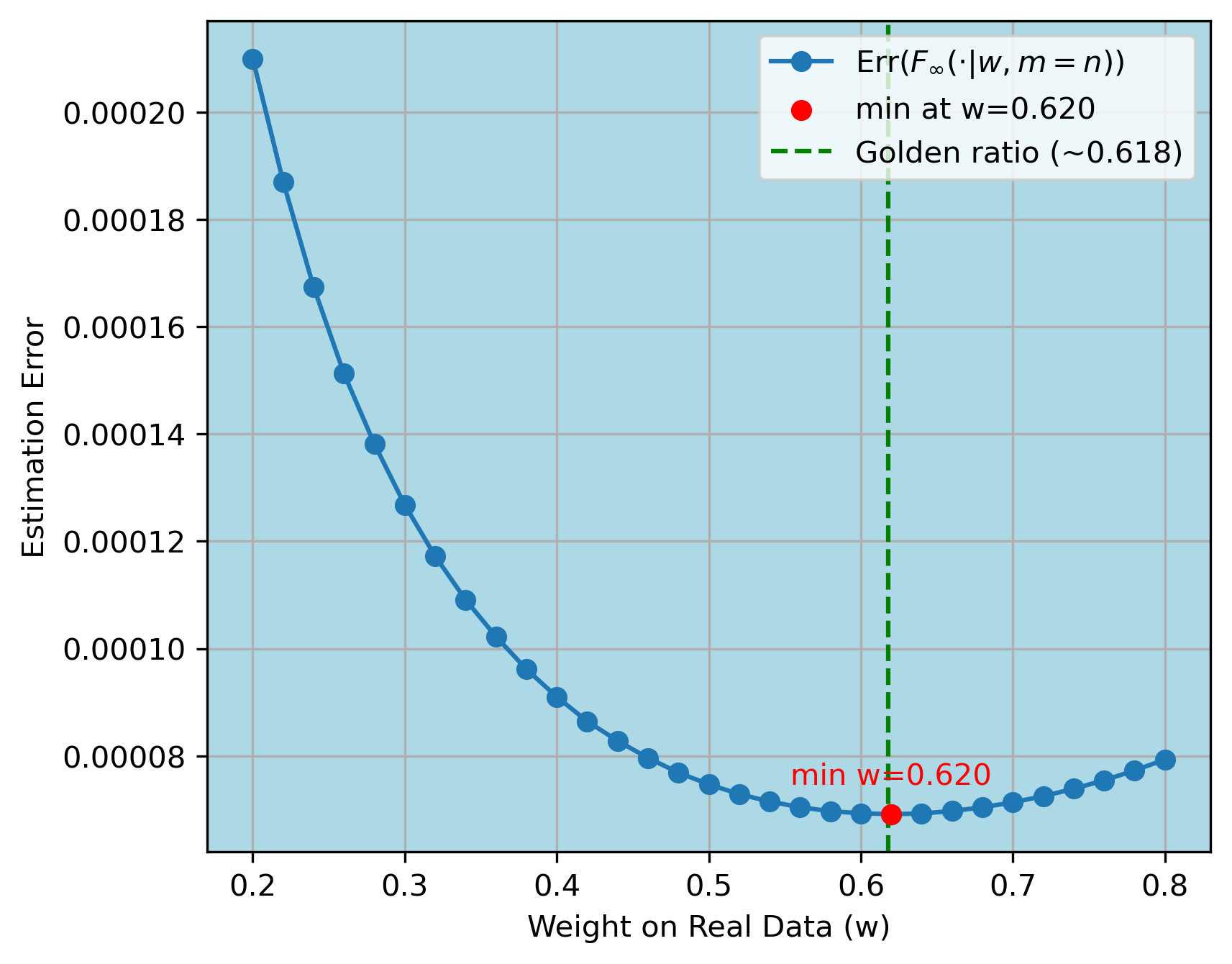}
        \caption{Workclass}
    \end{subfigure}
    \begin{subfigure}[b]{0.464\textwidth}
        \centering
        \includegraphics[width=\textwidth]{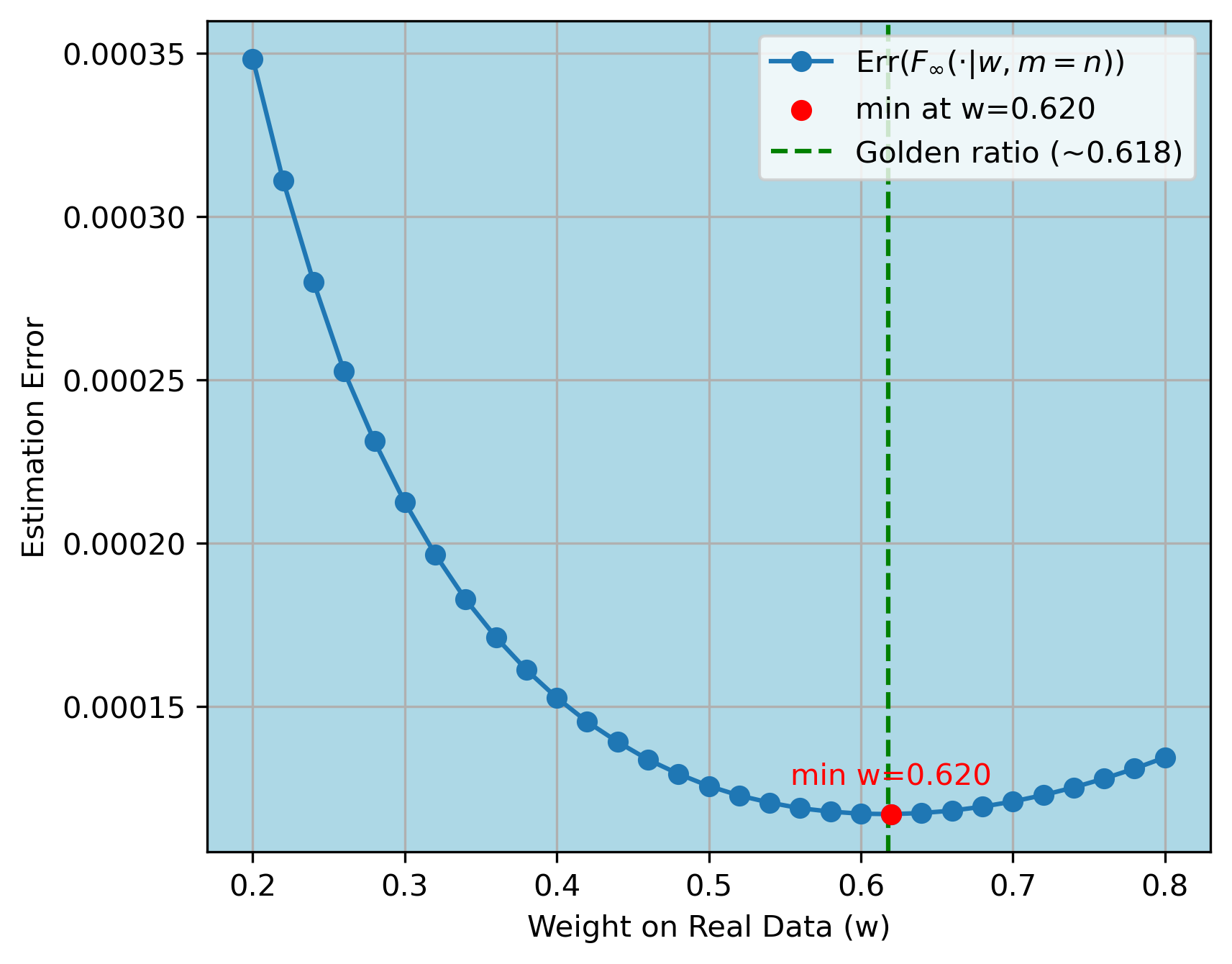}
        \caption{Marital Status}
    \end{subfigure}
        \begin{subfigure}[b]{0.464\textwidth}
        \centering
        \includegraphics[scale=0.523]{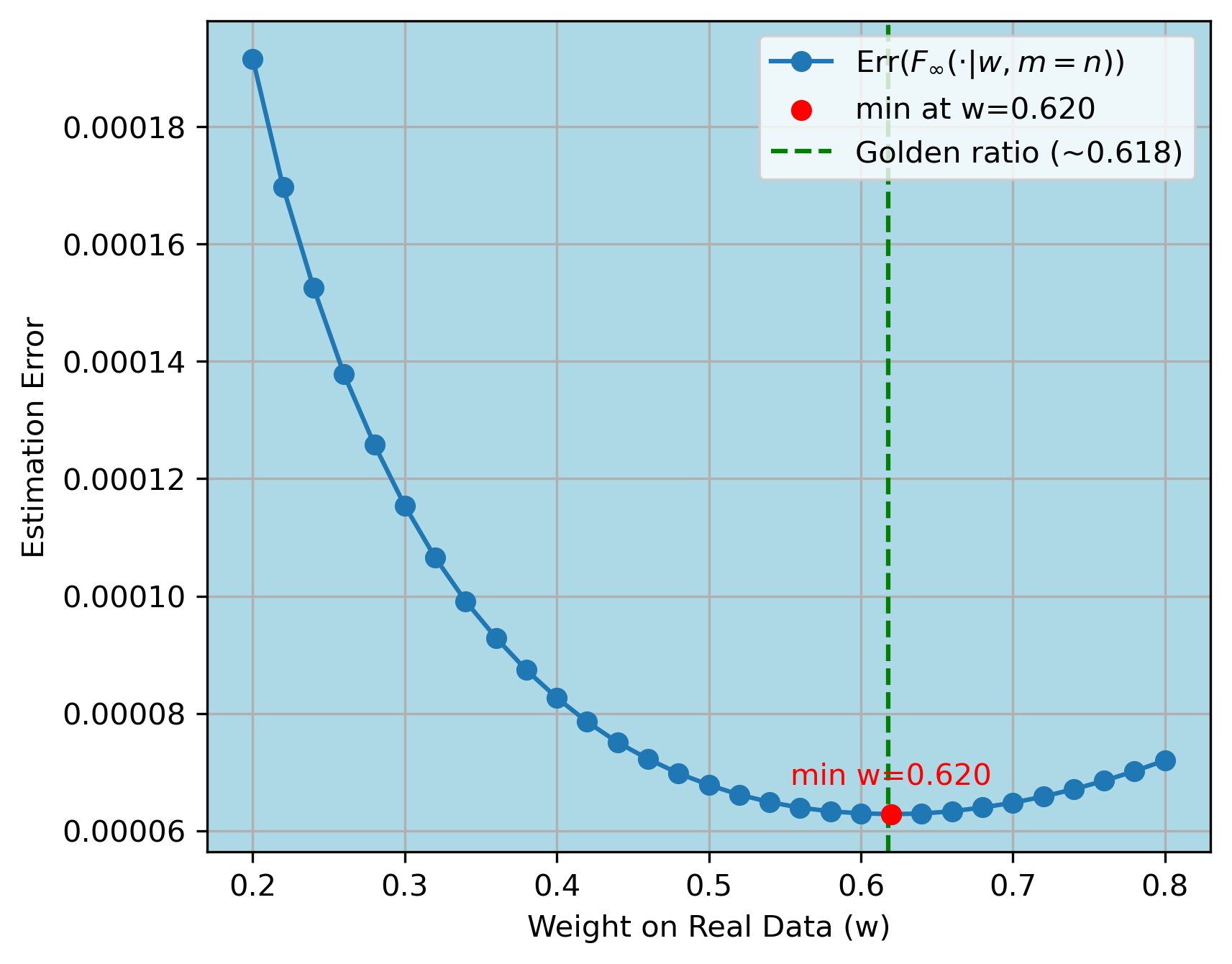}
        \caption{Education}
    \end{subfigure}
        \begin{subfigure}[b]{0.464\textwidth}
        \centering
        \includegraphics[scale=0.523]{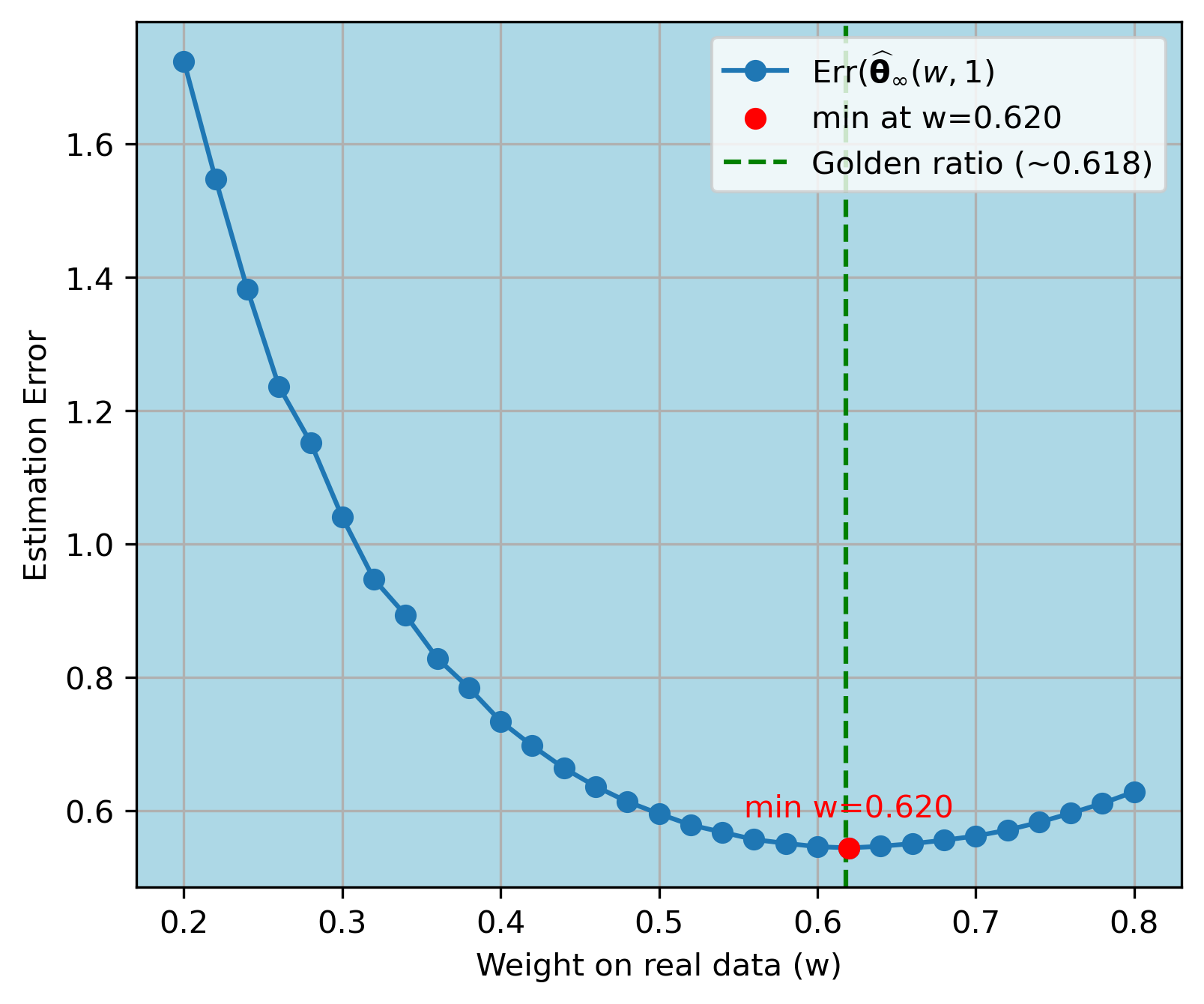}
        \caption{Logistic Regression}
    \end{subfigure}
    \caption{The averaged estimation errors of the proposed recursive weighted training scheme with $m = n = 1{,}000$ across different values of $w \in \{0.2 + 0.02 \times i : 0 \leq i \leq 30\}$.}
    \label{fig:real}
\end{figure}

To assess the benefits of recursive training with synthetic data, we first fit a logistic regression model on the entire dataset to obtain a pseudo ``ground-truth'' parameter vector $\bm{\theta}^\star$. We then conduct experiments by fixing the number of real and synthetic samples per iteration at $n = m = 500$ and performing $T = 100$ recursive updates. By varying the weight on real data $w$ from $0.2$ to $0.8$ in increments of $0.02$, we investigate its impact on the final estimation accuracy. For the nonparametric distribution estimation task, we similarly fix $n = m = 500$ and $T = 100$, and study the effect of $w$ over the same range.

In Figure \ref{fig:real}, the results indicate that the estimation errors across all cases reach their minimum near the reciprocal of the \textit{golden ratio}, implying that the optimal weight assigned to real data in recursive weighted training is approximately the reciprocal of the golden ratio.

\section{Conclusion and Future Work}
In this paper, we developed a theoretical framework to analyze model collapse under
recursive training with both synthetic and fresh real samples, and established optimal weighting strategies under limited data retention. Beyond analyzing how fresh real data can prevent model collapse, our results also
provide a quantitative characterization of the trade-offs between statistical efficiency
and memory constraints.

A key limitation of our analysis is its reliance on the ability to distinguish real and
synthetic samples, which may be imperfect in practice, especially when training data
are collected from large-scale web sources and lack reliable provenance information.

Future work includes extending the framework to settings with noisy or unavailable source information. In such settings, the weighting scheme may need to account for uncertainty about whether each sample is real or synthetic. Another important direction is to incorporate stronger identification mechanisms, such as improved statistical watermarking \citep{cai2024towards,li2024robust,li2025statistical,li2025optimal,xie2025debiasing} and related detection techniques. These extensions would help make the assumptions of the present framework more realistic in practical deployments.

\section*{Data Availability}
The Adult dataset is available from the 
\href{https://archive.ics.uci.edu/dataset/2/adult}{UCI Machine Learning Repository}.

\section*{Funding}
This work was supported in part by NSF grant CNS-2247795 and by Fundamental Research Funds for the Central Universities 20720261043.

\section*{Acknowledgement}
The authors are grateful to the Joint Editor, the Associate Editor, and the anonymous referees for their careful reading and constructive comments, which greatly improved the manuscript. We also gratefully acknowledge a gift from Cisco.

\section*{Conflicts of interest}
The authors do not declare any conflict of interest regarding this work.

\bibliographystyle{authoryear}
\putbib[ref]

\end{bibunit}

\end{document}